\documentclass[lettersize,journal]{IEEEtran}
\usepackage{amsmath,amsfonts}
\usepackage{algorithmic}
\usepackage{algorithm}
\usepackage[caption=false,font=normalsize,labelfont=sf,textfont=sf]{subfig}
\usepackage{textcomp}
\usepackage{tabularx}
\usepackage{stfloats}
\usepackage{url}
\usepackage{verbatim}
\usepackage{graphicx}
\usepackage{cite}
\usepackage{enumitem}
\usepackage{overpic}

\usepackage{mathrsfs}
\usepackage{amssymb}
\usepackage{eucal}
\usepackage{booktabs}   
\usepackage{makecell}
\usepackage{array}
\usepackage{xurl}
\usepackage{siunitx} 
\usepackage[colorlinks,urlcolor=blue,linkcolor=blue,citecolor=blue]{hyperref}
\makeatletter
\def\ps@pprintTitle{%
 \let\@oddhead\@empty
 \let\@evenhead\@empty
 \let\@oddfoot\@empty
 \let\@evenfoot\@empty
}
\makeatother

\begin{document}

\title{Are Kolmogorov Arnold Networks (KANs) Suitable for Imbalanced Data Classification?%
\author{
    Pankaj Yadav and Vivek Vijay}
\thanks{Reproducibility: Source code and full benchmark results are available at: \url{https://github.com/rogue816/KAN-Imbalanced}}%
\thanks{© 2026 IEEE. Personal use of this material is permitted. Permission from IEEE must be obtained for all other uses, in any current or future media, including reprinting/republishing this material for advertising or promotional purposes, creating new collective works, for resale or redistribution to servers or lists, or reuse of any copyrighted component of this work in other works. Accepted in \textit{IEEE Transactions on Emerging Topics in Computational Intelligence} (with supplementary material). DOI: \url{https://doi.org/10.1109/TETCI.2026.3747841}}
}

\maketitle

\begin{abstract}

Kolmogorov-Arnold Networks (KANs) parameterize each network connection as a learnable univariate B-spline, offering native feature-level interpretability and local representational flexibility. Whether these properties are assets or liabilities under class imbalance (where minority-class sparsity may leave individual spline edges without sufficient local observations to update reliably) remains unexamined. This study addresses that question by evaluating KANs against capacity-matched Multilayer Perceptrons (MLPs) and an XGBoost ensemble across ten imbalanced tabular datasets. This assessment covers three conditions: native class distribution, Focal Loss, and SMOTE-Tomek resampling, within a parameter-parity, two-phase experimental framework. Under the native distribution, KAN spline activations collapse in minority-sparse feature regions, exhibiting a more severe majority-class bias . Furthermore, Focal Loss yields only negligible improvement, as it reweights gradient contributions without supplying the localized minority-class observations. As a result, individual edge functions remain unable to develop discriminative curvature. In contrast, SMOTE-Tomek resampling enables spline activations to recover structured minority-class representations, achieving the best cohort-wide rank with a large and statistically significant effect. KANs native spline interpretability allows direct observation of these conditions through learned activation profiles. This simplifies explainability by providing direct evidence of representational change rather than threshold recalibration, avoiding the post-hoc attributions. However, these interpretability gains come at a steep computational cost. Across all evaluated conditions, KANs training incurs substantially higher overhead than a parameter-matched MLP, showing no strategy-dependent reduction in training latency. These findings characterize KANs as conditionally suitable for imbalanced tabular classification: effective and interpretable when data-level resampling is applied, and costly relative to simpler alternatives when it is not.

\end{abstract}


\section{Introduction}
\label{sec:introduction}

Class-imbalanced learning remains a structural and critical property in supervised pattern recognition~\cite{kan1, kan41}. The minority instances representing rare events such as fraud, disease onset, or hardware faults are systematically outnumbered by majority-class observations~\cite{kan43, kan42}. This imbalance frequently degrades classification performance by biasing the standard optimization process toward the majority class. While foundational in tabular settings~\cite{kan36}, data sparsity and highly skewed distributions present severe challenges across diverse domains, including signal propagation dynamics and operational resilience within complex topological networks~\cite{kan29,kan30}. In standard tabular environments, this problem is compounded by feature-space sparsity in the minority region. The standard cross-entropy training concentrates gradient signal on the dense majority class, leaving classifiers that perform well on aggregate accuracy metrics while failing entirely on the minority class~\cite{kan2}. Established strategies for addressing this challenge fall into two families. Data-level interventions (synthetic over-sampling, boundary filtering) augment the training distribution directly~\cite{kan22, kan37}. Loss-level interventions (class weighting, Focal Loss) redirect gradient contributions without altering the training data~\cite{kan20}. Both families are well-studied, yet their interaction with specific architectural mechanisms remains system-dependent and incompletely characterized~\cite{kan49}.

Multilayer Perceptrons (MLPs) address imbalanced classification through globally coupled weight matrices and fixed, predefined activation functions. This design provides consistent gradient flow across the full training distribution, giving fixed-basis classifiers a degree of robustness to minority-class sparsity~\cite{kan3,kan40}. The cost is representational opacity as weight matrices encode feature contributions in a distributed, entangled form that cannot be inspected directly. Therefore, these specific architectures require post-hoc attribution methods such as SHAP or LIME to relate internal representations to individual inputs~\cite{kan38}. In risk-sensitive real-world deployments, this opacity limits the diagnostic utility of the model even when its predictive accuracy is adequate. This limitation is particularly critical in fields such as clinical diagnostics, industrial fault detection, financial risk scoring, and critical infrastructure frameworks like emergency vehicle trajectory prediction or zero-trust security computing~\cite{kan34,kan44,kan35}. Beyond structured data infrastructures, a parallel demand for robust, interpretable modeling exists in advanced computer vision systems. This need is acutely observed in tasks like unsupervised domain adaptation for nighttime semantic segmentation and low-light salient object detection~\cite{kan31, kan32, kan33}. In these environments, severe distribution shifts and extreme lighting constraints heavily degrade standard deep learning representations, making model transparency essential for reliable deployment. 

Kolmogorov-Arnold Networks (KANs) present a structurally distinct alternative. Grounded in the Kolmogorov-Arnold Representation Theorem~\cite{kan4, kan5}, KANs replace the fixed activation and weight-matrix design of MLPs with learnable univariate B-spline functions placed on each network edge. Each edge function $\phi_{i,j}(x)$ is a piecewise polynomial whose coefficients are optimized during training, adapting locally to the distribution of observations that activate a specific edge~\cite{kan45,kan47}. This local adaptivity is precisely the property that makes KANs interpretable. Each $\phi_{i,j}(x)$ is a complete, readable transformation of a single input feature, inspectable as a curve without approximation or attribution~\cite{kan6}. But local adaptivity is also the property that raises a non-trivial concern under class imbalance. If a feature's input range is sparsely populated in the minority-class region, the corresponding spline coefficients may receive insufficient gradient signal to develop meaningful curvature. Consequently, these activation profiles default to flat, majority-dominated curves that contribute little to minority-class discrimination.
 
This concern remains unaddressed in the current literature. Existing studies focus primarily on physics-informed scientific computing, function approximation, and partial differential equations~\cite{kan7, kan8}. These tasks are characterized by dense, uniform input coverage, which contrasts sharply with the skewed, sparse distributions inherent to imbalanced classification. The KANBalance framework~\cite{kan39} has explored loss-level modifications to rebalance spline gradient updates. However, that work lacks a data-level comparison or a parameter-matched neural baseline, leaving open the question of whether the spline mechanism itself is the limiting factor or whether standard imbalance treatments suffice.

\begin{figure}[!htbp]
\centering
\begin{overpic}[width=\columnwidth]{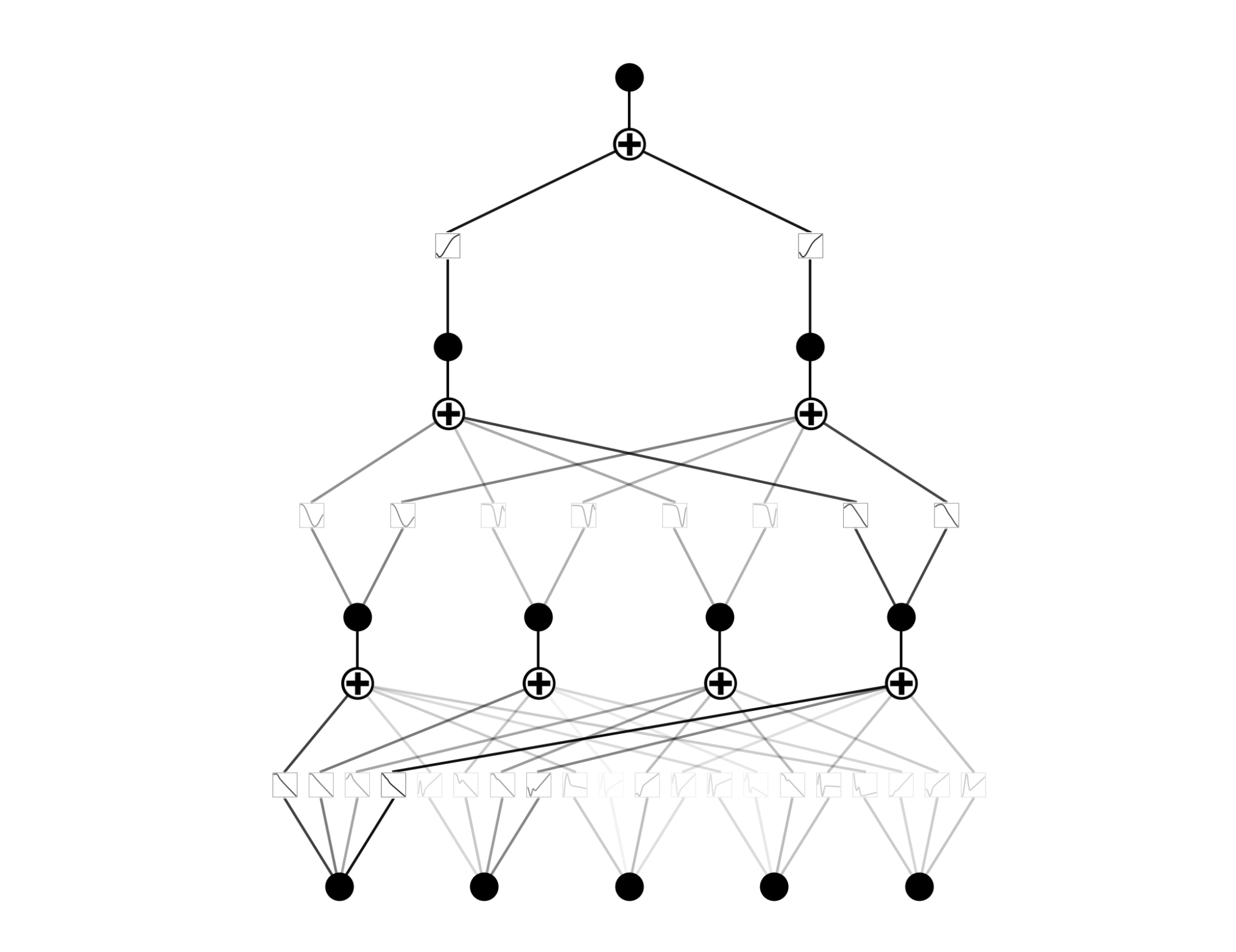}
    
    \put(80,40){\rotatebox{-90}{\scriptsize Network Depth ($L$)}}
    
    \put(30,1){\scriptsize Network Width $[n_0, n_1, \dots, n_l]$}
    
    
    \put(6,43){\scriptsize Summation Nodes ($\sum$)}
    
    \put(0.5,12){\scriptsize\shortstack[l]{Edge Activation\\ Functions ($\phi_{i,j}$)}}
\end{overpic}
\caption{Structural topology of a KAN, showing edge-based learnable univariate B-spline activation functions ($\phi_{l,i,j}$) connecting summation layers ($\sum$) across network width ($n_l$) and depth ($L$). Each edge carries its own trainable spline function rather than a scalar weight.}
\label{fig:arch}
\end{figure}

This study addresses that gap directly. We ask: \textit{Are Kolmogorov-Arnold Networks Suitable for Imbalanced Data Classification?} To answer it, we design a two-phase empirical framework. Phase~I calibrates the spline grid resolution through a multi-seed sensitivity sweep on two datasets spanning the moderate and severe ends of the imbalance spectrum. Phase~II applies the calibrated configuration to a ten-dataset benchmark, comparing KANs against capacity-matched MLPs and XGBoost. These models are evaluated under three conditions: native distribution (baseline), Focal Loss, and SMOTE-Tomek resampling. The KAN architecture, with its edge-based learnable activation functions, is illustrated in Fig.~\ref{fig:arch}.


The contributions of this work are:
\begin{enumerate}[label=(\roman*), leftmargin=*, noitemsep, topsep=2pt]
    \item \textit{Controlled Parameter-parity benchmark:} We evaluate KANs against XGBoost and parameter-matched MLPs across ten imbalanced KEEL datasets. By analytically deriving the MLP hidden width to maintain a strict parameter parity ratio ($\approx 1.0$), this experimental design successfully isolates internal architecture behaviors from raw model capacity differences.
    \item \textit{Quantified interaction between spline mechanics and imbalance treatment:} We demonstrate that data-level resampling (SMOTE-Tomek) yields substantial, statistically significant improvement in KAN minority-class performance over loss-level reweighting (Focal Loss). Our findings demonstrate that the localized nature of KAN edge activations make them uniquely dependent on the local data coverage provided by distributional augmentation.
    \item \textit{Characterization of KAN training overhead:} We quantify that KAN training incurs a consistent $30$--$43\times$ computation slowdown compared to parameter-matched MLPs across all evaluated cohorts. This overhead is fundamentally architectural, originating from sequential B-spline coefficient evaluations rather than data or loss-level imbalance mitigation strategies. 
    \item \textit{Mechanistic case study through direct activation inspection:} We examine the learned $\phi_{i,j}(x)$ profiles under baseline and resampled conditions, revealing that natively suppressed edge functions develop structured non-monotone curvature after resampling. This provides direct visual evidence of representational shifts at the edge level, highlighting the native interpretability of KANs in settings where MLP require complex post-hoc attribution methods. 
\end{enumerate}

The remainder of this article is organized as follows. Section~\ref{sec:2} introduces the mathematical formulation of KANs. Section~\ref{sec:3} details the experimental methodology. Section~\ref{sec:results} reports the empirical findings. Section~\ref{sec:discussion} presents the activation analysis and deployment characterization. Section~\ref{sec:conclusion} concludes the study.

\section{Kolmogorov Arnold Networks (KANs)}
\label{sec:2}

The section provides  the theoretical and architectural foundations of KANs. It first introduces the Kolmogorov-Arnold Representation Theorem, which establishes the mathematical basis for the architecture. It then details the KAN architecture, including its spline-based functional parameterization and resulting interpretability properties. Finally, it extends the discussion to deep Multi-KAN architectures and the implementation mechanics relevant to practical deployment.

\subsection{Kolmogorov Arnold Representation Theorem}
\label{subsec:1}
Kolmogorov Arnold Networks (KANs) derive their theoretical foundation from the Kolmogorov-Arnold Representation Theorem, established by Andrey Kolmogorov and Vladimir Arnold in 1957~\cite{kan9}. This result states that \emph{any continuous multi-variate function} defined on a bounded domain can be represented as a finite superposition of continuous univariate functions and addition.  

Formally, let $f(\mathbf{x}) : [0,1]^n \rightarrow \mathbb{R}$ be an arbitrary continuous \emph{multivariate} function of $n$ inputs where $\mathbf{x} = (x_1, x_2, \dots, x_n)^T$. There exist continuous univariate function $\phi_{ij}: [0,1] \rightarrow \mathbb{R}$ and $\psi_i : \mathbb{R} \rightarrow \mathbb{R}$ such that ~\cite{kan10}:

\begin{equation}
     f(\mathbf{x}) = \sum_{i=1}^{2n+1}\psi_i \left(\sum_{j=1}^{n} \phi_{ij}{(x_j)} \right),
    \label{eq:kan1}
\end{equation}

for all $\mathbf{x} \in [0,1]^n$. The significance of this decomposition is that the $n$-dimensional function $f$ is reduced entirely to composition of one dimensional function. This means there are no inherently multivariate operations. The inner function $\phi_{ij}$ transform each scalar input $x_j$ independently, and the outer function $\psi_i$ act only on the scalar aggregations $\zeta_i = \sum_{j=1}^n \phi_{ij}(x_j)$. Multivariate complexity is thus eliminated in the sense that all function approximation is delegated to univariate mappings and their additive composition. 

This decomposition reveals a two-layer feedforward structure: an \emph{inner transformation layer} that evaluates $n(2n+1)$ univariate functions $\phi_{ij}(x_j)$, followed by $2n+1$ linear aggregations $\zeta_i$, and an \emph{outer transformation layer} that applies a distinct univariate mapping $\psi_i(\cdot)$ to each aggregated scalar.


The theorem achieves substantial dimensionality reduction by projecting high-dimensional inputs onto scalar representations via the inner aggregations $\zeta_i$~\cite{kan11}. The bounded outer width of $2n+1$ was proven optimal by Arnold, establishing that multivariate continuity can always be expressed through a finite composition of univariate operations. This representation is closely related to Hilbert's thirteenth problem, showing that multivariate functions can be evaluated without inherently multi-dimensional operations~\cite{kan12}. In machine learning, this framework offers three primary advantages: 1)~existence guarantees on compact sets, 2)~operational simplicity through the elimination of multivariate convolutions, and 3)~adaptive structural design, where deep KAN architectures relax the strict $2n+1$ constraint in favor of hierarchical feature extraction. 

\subsection{KAN Architecture and Spline Approximation Foundations}
\label{subsec:2}

While the Kolmogorov-Arnold theorem guarantees the existence of univariate functions, it provides no mechanism for parameterizing them. Translating the theorem into a trainable neural architecture requires a robust function approximation scheme. Standard global polynomial representations, such as high-degree Weierstrass or Taylor expansions, are susceptible to the \textbf{Runge Phenomenon}, in which polynomial approximations exhibit oscillatory divergence near interval boundaries when fitted to discrete data~\cite{kan13}. 

Formally, for a target univariate continuous function $h:[a,b] \rightarrow \mathbb{R}$ approximated by a $p$-degree global polynomial $P_p(x) = \sum_{s=0}^p c_sx^s$, the uniform approximation error is bounded by: 

\begin{equation}
\sup_{x \in [a,b]} |h(x) -P_p(x)| \leq \frac{(b-a)^{p+1}}{(p+1)!} \sup_{\xi \in [a,b]} |h^{(p+1)}(\xi)|,
\end{equation}

For complex, real-world data distributions, the high-order derivatives $h^{(p+1)}(\xi)$ scale exponentially, leading to numerical instability. To address this, early approximation frameworks introduced B\'{e}zier curves. A B\'{e}zier curve of degree $p$, parameterized by control points $\mathbf{c} = (c_0, c_1, \dots, c_p)^T \in \mathbb{R}^{p+1}$, is defined via Bernstein basis polynomials:

\begin{equation}
B_p(x; \mathbf{c}) = \sum_{s=0}^p c_s \binom{p}{s}x^s(1-x)^{p-s}.
\end{equation}

Although B\'{e}zier curves guarantee strict convex hull confinement and exact endpoint interpolation, adjusting a single control point $c_s$ alters the entire curve across the domain $[0,1]$. Consequently, for degrees $p>5$, B\'{e}zier curves become more prone to numerical instability and oscillatory behavior~\cite{kan14}.

To achieve localized control and numerical stability, KANs replace global polynomials with \emph{B-splines}. The domain $[0,1]$ is partitioned into $G$ uniform sub-intervals via a base grid of $G+1$ interior knots $\{t_1, \dots, t_{G+1}\}$. To support B-spline basis functions of degree $k$ at the domain boundaries, the base grid is extended by repeating the boundary value $k$ times on each side, yielding the \emph{extended knot vector}:

\begin{equation}
\begin{split}
    \boldsymbol{\tau} = \Big\{ & \underbrace{t_1 = \dots = t_k}_{k \text{ copies}} = 0 \\
    & < t_{k+1} < \dots < t_{G+k} \\
    & < 1 = \underbrace{t_{G+k+1} = \dots = t_{G+2k}}_{k \text{ copies}} \Big\}.
\end{split}
\end{equation}

The extended knot vector contains $G+2k$ elements (indices $1$ to $G+2k$). On this extended knot vector, the number of B-splines basis functions of degree $k$ is: 
\begin{equation}
    N = G + k
    \label{eq:N_basis}
\end{equation}

This is a standard result: on a knot sequence of $M$ elements, the number of B-spline basis functions of degree $k$ is $M - k - 1$, so $(G+2k) - k - 1 + 1 = G + k$ with 1-based indexing. A B-spline function $S(x; \boldsymbol{\theta})$ of degree $k$ with $N = G + k$ basis functions is defined as:

\begin{equation}
S(x; \boldsymbol{\theta}) = \sum_{m=1}^{G+k} \theta_m B_{m,k}(x),
\end{equation}

where, $\boldsymbol{\theta} \in \mathbb{R}^{G+k}$ is the vector of trainable spline coefficients, and the basis functions $B_{m,k}(x)$ are evaluated recursively using the Cox-de Boor recursion~\cite{kan15}. With $k$ interpreted as the spline \emph{degree} (so cubic splines have $k=3$), the degree-0 base
case and the degree-$k$ recursion are:

\begin{equation}
B_{m,0}(x) = \begin{cases} 1, & t_m \leq x < t_{m+1} \\ 0, & \text{otherwise} \end{cases}
\label{eq:cdb_base}
\end{equation}
\begin{equation}
\begin{split}
B_{m,k}(x) = \frac{x - t_m}{t_{m+k} - t_m} B_{m,k-1}(x)
+ \\
\frac{t_{m+k+1} - x}{t_{m+k+1} - t_{m+1}} B_{m+1,k-1}(x)
\end{split}
\label{eq:cdb_recursion}
\end{equation}

B-splines offer three distinct advantages over global polynomials and  B\'{e}zier representations:
\begin{enumerate}
    
\item \textbf{Compact local support:} $\text{supp}(B_{m,k}) = [t_m, t_{m+k+1}]$, so updates to $\theta_m$ affect only a local neighborhood of the input domain. 

\item \textbf{High-order continuity:} $S(x) \in C^{k-1}([0,1])$ at simple interior knots, ensuring smooth gradient flow during optimization. For the cubic splines ($k=3$) used throughout this study, this gives $C^2$ continuity.

\item \textbf{Convex hull confinement:} For any $x \in [t_m, t_{m+1}]$, $S(x) \in [\min_m \theta_m, \max_m \theta_m]$, bounding the function's output range~\cite{kan16}.

\end{enumerate}

In the basic KAN formulation, each weight connection is replaced by an independent univariate B-spline function:

\begin{equation}
\phi_{ji}^{(l)}(x) = S_{ji}^{(l)}(x; \boldsymbol{\theta}_{ji}^{(l)}),
\end{equation}

where  $\boldsymbol{\theta}_{ji}^{(l)} \in \mathbb{R}^{N}$ denotes the trainable spline coefficients connecting input node $i$ to output node $j$ within layer $l$. In practice, this pure B-spline parameterization is augmented with a residual base term to stabilize training, as described in~\ref{subsec:3}

\subsection{Deep Multi-KAN Composition and Implementation Mechanics}
\label{subsec:3}

The Kolmogorov-Arnold formulation in Equation~\eqref{eq:kan1} defines a two layer structure ($L=2$): one inner layer applying $\mathbf{\phi_{ij}}$ functions and one outer layer applying the $\psi_{i}$ functions. The Multi-KAN architecture generalizes this to deep compositions by stacking $L \geq 2$ layers of spline-parameterized univariate functions, forming a hierarchical computational graph~\cite{kan17}. For an $L$-layer Multi-KAN with width configuration $[d_0, d_1, \dots, d_L]$, where $d_{0}=n$ is the input dimension, $d_{L}=D_{out}$ is the output dimension, and $d_{l}$ denotes the number of nodes in layer $l$, the forward pass is:

\begin{equation}
f(\mathbf{x}) = \mathbf{\Phi}^{(L)} \circ \mathbf{\Phi}^{(L-1)} \circ \dots \circ \mathbf{\Phi}^{(1)}(\mathbf{x}),
\label{eq:kan2}
\end{equation}

where each transformation $\mathbf{\Phi}^{(l)} : \mathbb{R}^{d_l} \rightarrow \mathbb{R}^{d_{l+1}}$ is evaluated component-wise. Setting $\mathbf{z}^{(0)} = \mathbf{x}$:

\begin{equation}
\mathbf{z}^{(l+1)} = \mathbf{\Phi}^{(l)}(\mathbf{z}^{(l)}) = \left[ \sum_{i=1}^{d_l} \phi_{ji}^{(l)}(z_i^{(l)}) \right]_{j=1}^{d_{l+1}}.
\end{equation}

\paragraph{Residual spline parameterization} 
To stabilize gradient propagation across deep compositions and preserve a linear transformation pathway, practical Multi-KANs implementations replace the pure B-spline edge function with a two-term residual activation~\cite{kan18}.

\begin{equation}
\begin{split}
\phi_{ji}^{(l)}(x) = w_b \cdot b(x) + w_s \cdot \text{spline}(x) =\\
w_b \cdot \text{SiLU}(x) + w_s \cdot \sum_{m=1}^{G+k} \theta_m B_{m,k}(x),
\end{split}
\end{equation}

where, $b(x) = \text{SiLU}(x) = \frac{x}{1 + e^{-x}}$ is a smooth, globally differentiable base activation that providing a linear residual pathway, and $w_{b}$, $w_{s}$ are trainable scalar coefficients balancing the global base term against the local spline term. This parameterization helps prevent vanishing or exploding gradients. Therefore, the SiLU term adapts to local variation, while the spline term captures local structure. The derivative with respect to the input decomposes as:

\begin{equation}
\begin{split}
\frac{d\phi_{ji}^{(l)}}{dx}
= {} & w_b \cdot \frac{1 + e^{-x} + x e^{-x}}{(1 + e^{-x})^2} \\
& + w_s \sum_{m=1}^{G+k} \theta_m \left(
    \frac{k}{\,t_{m+k} - t_m\,} B_{m,k-1}(x) \right. \\
   &\left. - \frac{k}{\,t_{m+k+1} - t_{m+1}\,} B_{m+1,k-1}(x)
  \right).
\end{split}
\end{equation}

\paragraph{Universal approximation guarantees} 
Deep Multi-KANs retain universal approximation guarantees. For any continuous $f : [0,1]^n \to \mathbb{R}$ and $\epsilon > 0$, there exists a Multi-KAN of depth $L$ and widths $d_l$ such that:

\begin{equation}
\sup_{\mathbf{x} \in [0,1]^n} |f(\mathbf{x}) - \text{Multi-KAN}(\mathbf{x})| < \epsilon.
\end{equation}

For a $q$-times continuously differentiable target function, KAN models achieve an approximation error convergence rate of $\mathcal{O}(G^{-q})$ under uniform grid refinement~\cite{kan19}. Because the total trainable parameter count scales as $N_p = (G + k)\cdot N_{\text{edges}}$ (plus a small constant per edge from the residual terms $w_b$ and $w_s$), increasing $G$ adds exactly one spline coefficient per network edge. This linear growth is the parameter scaling trajectory observed empirically in the Phase~I sensitivity sweep (Section~\ref{subsec:phase1_results}).

However, when applied to real-world tabular data under severe class imbalance, the compact support property introduces a specific limitation. Because each coefficient $\theta_m$ is updated only by training
samples falling within $\text{supp}(B_{m,k}) = [t_m, t_{m+k+1}]$, increasing $G$ produces narrower support intervals. When minority-class samples are sparse, many of these intervals contain no minority observations, so the corresponding coefficients remain uninformative regardless of grid resolution. This motivates the grid sensitivity analysis introduced in Section~\ref{subsec:grid_sensitivity} and reported in Section~\ref{subsec:phase1_results}, which examines the trade-off between grid resolution and robustness to minority-class sparsity.

\section{Methodology}
\label{sec:3}

This study evaluates the structural stability, classification performance, and computational cost of Kolmogorov-Arnold Networks (KANs) under varying degrees of class imbalance through a controlled, parameter-matched experimental framework. The pipeline proceeds in two sequential phases. An initial spline calibration phase (\textbf{Phase~I}), which sweeps a discrete grid resolution spectrum over multiple random seeds to determine a stable baseline spline configuration. Followed by a comparative benchmarking phase (\textbf{Phase~II}), which evaluates the calibrated KAN against parameter-matched neural and ensemble baselines across ten imbalanced datasets. 
 The subsequent subsections describe each component. Section~\ref{subsec:empirical_profile} describes the dataset cohort and hardware environment. Section~\ref{subsec:grid_sensitivity} details the Phase~I spline calibration, including the parameter-matching procedure. Section~\ref{subsec:phase2_benchmark} describes the Phase~II benchmarking setup, including baseline architectures. Section~\ref{subsec:interventions} specifics the imbalance mitigation strategies, and Section~\ref{subsec:statistical_validation} defines the evaluation metrics and statistical validation procedures.

\subsection{Empirical Data Landscapes and Hardware Environment}
\label{subsec:empirical_profile}

 The experiment spans a benchmark cohort of ten  small-to medium-sized imbalanced datasets drawn from the Knowledge Extraction based on Evolutionary Learning (KEEL) repository~\cite{kan21}. Because KANs parametrize activations as learnable B-splines, their updates rely on localized univariate function evaluations rather than standard, monolithic matrix multiplications. This edge-based optimization introduces a known theoretical computational overhead.  To accommodate this constraint, a ten-dataset cohort was selected to provide a sufficiently diverse evaluation scope while remaining computationally tractable.

The selected datasets span a range of diverse range of problem characteristics. With input dimensionality from $d_{\text{in}}= 5$ to $11$, sample sizes from $N=214$ to $1484$ and imbalance ratios (IR) from $5.14$ to $46.50$. All instances represent natively binary classification tasks. Categorical attributes are ordinally encoded, and the minority class is assigned the positive label ($y=1$) throughout. Full dataset characteristics are provided in Table~\ref{tab:data}.

\begin{table}[!htbp]
\centering
\caption{ Characteristics of the benchmarking dataset cohort}
\label{tab:data}
\scriptsize
\renewcommand{\arraystretch}{1.3}
\begin{tabular}{lccc}
\toprule
\textbf{Dataset} & \textbf{Features ($d_{\text{in}}$)} & \textbf{Instances ($N$)} & \textbf{Imb. Ratio ($\text{IR}$)}  \\
\midrule
yeast4 & 8 & 1484 & 28.10  \\
yeast5 & 8 & 1484 & 32.73 \\
yeast6 & 8 & 1484 & 41.40  \\
glass2 & 9 & 214 & 11.59 \\
ecoli3 & 7 & 336 & 8.60  \\
winequality-red-8\_vs\_6-7 & 11 & 855 & 46.50 \\
new-thyroid1 & 5 & 215 & 5.14 \\
glass4 & 9 & 214 & 15.47  \\
glass6 & 9 & 214 & 6.02  \\
winequality-red-8\_vs\_6 & 11 & 656 & 35.44 \\
\bottomrule
\end{tabular}
\end{table}

To ensure reproducibility, all experiments (including Phase~I and Phase~II) are executed on a fixed, dedicated hardware environment. All computations are performed on an NVIDIA A100 Tensor Core GPU configured as a Multi-Instance GPU (MIG) slice (3g.20gb profile), providing 20~GB of dedicated HBM2e memory on the Ampere (GA100) architecture with approximately 1,728 CUDA cores, 108 third-generation Tensor Cores, and CUDA 12.x. This fixed, isolated partition also ensures that runtime and memory measurements are not confounded by competing workloads.

\subsection{Phase I: Spline Grid Sensitivity Analysis}
\label{subsec:grid_sensitivity}

The goal of Phase~I is to analyze a spline grid resolution $G_{\text{grid}}$ that can yield stable minority-class performance before the full ten-dataset benchmark evaluated. The analysis is restricted to two datasets, \texttt{ecoli3} and \texttt{yeast5}. This restriction is motivated by the sequential nature of B-spline coefficient updates in current KAN implementations. It makes grid resolution sweeps across the full ten-dataset cohort computationally prohibitive. Consequently, a focused two-dataset subset makes the sweep tractable while preserving generalizability through the following design choices:

\begin{enumerate}[label=(\roman*), leftmargin=*, noitemsep, topsep=2pt]
    \item \textit{Imbalance range coverage:} \texttt{ecoli3} ($\text{IR} = 8.60$) represents a moderate imbalance setting and \texttt{yeast5} ($\text{IR} = 32.73$) represents a severe imbalance setting, capturing spline behavior across both ends of the cohort's imbalance range. 
    \item \textit{Dimensional proximity:} \texttt{ecoli3} ($d_{\text{in}} = 7$) and \texttt{yeast5} ($d_{\text{in}} = 8$) have closely matched input dimensions, so that differences in spline behavior across grid resolutions can be attributed to the grid density rather than to input space variation. 
\end{enumerate}

Phase I additionally restricts the imbalance mitigation strategy to SMOTE-Tomek resampling only, excluding  Focal Loss from the grid sweep. SMOTE-Tomek directly augments the training distribution by generating synthetic minority samples and removing borderline majority instances, making it's interaction with spline grid density more directly observable. Conversely, Focal Loss, modifies only the loss weighting without altering the underlying training distribution. The optimal grid configurations determined through this phase are subsequently carry forwarded to Phase II, where both SMOTE-Tomek and Focal Loss are evaluated across the full cohort. 
The Phase I optimization proceeds across three sequential steps:

\textbf{Step~1- Parameter-parity constrained architecture search.}\\
A strict parameter-parity constraint is enforced to ensure that performance differences between the KAN and MLP reflect structural differences rather than capacity differences, throughout both phases.

Given a KAN parameter budget $N_{\text{target}}$, the MLP hidden width $W_{\text{MLP}}$ is derived in closed form by equating total trainable parameters and solving for the hidden dimensions. For a single hidden layer ($n_{\text{layer}}=1$) with binary output ($D_{\text{out}}=2$):
\begin{equation}
    W_{\text{MLP}} = \max\left(8, \left\lfloor \frac{N_{\text{target}} - D_{\text{out}}}{d_{\text{in}} + D_{\text{out}} + 1} \right\rfloor\right)
    \label{eq:mlp_single}
\end{equation}

For two hidden layers of equal width ($n_{\text{layers}} = 2$), the parameter count yields a quadratic whose positive root gives:
\begin{equation}
\resizebox{\columnwidth}{!}{$
    W_{\text{MLP}} = \max\left(8, \left\lfloor \frac{-(d_{\text{in}} + D_{\text{out}} + 2) + \sqrt{(d_{\text{in}} + D_{\text{out}} + 2)^2 - 4(D_{\text{out}} - N_{\text{target}})}}{2} \right\rfloor\right)
    $}
    \label{eq:mlp_deep}
\end{equation}

The $\max(8, \cdot)$ floor prevents degenerate architectures when compact KAN configurations produce fractional widths below a meaningful threshold. Applied to Phase~I  datasets, this formulation yields parity ratios of $0.98$ and $0.97$ for \texttt{ecoli3} and \texttt{yeast5} respectively. Consequently, confirming that any performance gap is attributable to architectural behavior rather than parameter budget.

\textbf{Step~2- Hyperparameter optimization under fixed grid resolution.} Using the parity-constrained search space, an optuna based hyperparameter optimization (HPO) with 5-fold cross-validation targets the validation Geometric Mean ($G_{\text{mean}}$). To ensure strict parity, this HPO process is executed independently for the baseline and resampled configurations using identical random seeds. This grid resolution is held fixed at $G_{\text{grid}} = 5$ during the HPO, isolating it as an independent variable for subsequent sweep. Each independent optimization runs for twenty-five trials across the architectural spaces summarized in Table~\ref{tab:hpo_spaces}. The MLP width ($W_{\text{MLP}}$) is recomputed at each trial via \eqref{eq:mlp_single} or~\eqref{eq:mlp_deep} as the KAN architecture varies. The resulting optimized hyperparameter configurations for both strategies are subsequently frozen and transferred to the downstream grid resolution sweeps. 

\begin{table}[!htbp]
\centering
\caption{Hyperparameter Search Spaces for Phase~I optimization.}
\label{tab:hpo_spaces}
\scriptsize
\renewcommand{\arraystretch}{1.2}
\begin{tabularx}{\columnwidth}{llX}
\toprule
\textbf{Group} & \textbf{Parameter} & \textbf{Search Space / Setting} \\
\midrule
\textit{Optimization}   & Learning Rate ($lr$) & $[10^{-4}, 5\times 10^{-3}]$ (log-Uniform) \\
                               & Weight Decay ($\lambda$)  & $[10^{-5}, 10^{-2}]$ (log-Uniform) \\
                               & Batch Size ($\mathcal{B}$) & $\{16, 32\}$ \\
                               & Trials & 25 (fixed) \\
\midrule
\textit{KAN geometry}          & Hidden depth ($L$)    & $\{1, 2\}$ \\
                               & Node width ($w_i$)   & $[2, 12]$  \\
                               & Spline degree ($k$)  & $\{2, 3\}$ \\
                              
\midrule
\textit{MLP baseline}     & Width ($W_{\text{MLP}}$) & Derived via \eqref{eq:mlp_single} or \eqref{eq:mlp_deep} \\
\midrule
\textit{Sensitivity Sweep}    & Grid resolution & $G_{\text{grid}} \in \{3, 5, 7, 9, 12\}$ \\
                               & Random seeds ($\mathcal{S}$) & $\{42, 123, 999, 7, 21\}$ \\
\bottomrule
\end{tabularx}
\end{table}

\textbf{Step~3- Isolated grid resolution sweep.} All architectural and optimization parameters identified in Step~2 are frozen. The grid resolution is then varied as the sole free parameter across the candidate values in Table~\ref{tab:hpo_spaces}, yielding 25 evaluation runs per training condition. The multi-seed evaluation controls for initialization variance, allowing the effect of grid resolution on minority-class classification performance to be assessed independently of random initialization. 

The grid resolution identified from this analysis is transferred as a stable global configuration to the Phase~II benchmark. Framing this as an inference from an observed stability plateau (rather than a fixed assignment) is appropriate because the sweep exposes an empirical performance curve rather than a prescribed optimum. 

\subsection{Phase II: Comparative Benchmarking}
\label{subsec:phase2_benchmark}

Following the spline calibration conducted in Phase I, the evaluation framework moves to the main comparative benchmark across the full ten-dataset cohort. This phase benchmarks the calibrated KAN against a parameter-matched MLP and an XGBoost ensemble under uniform experimental conditions. To control for stochastic variability, all models are evaluated using a multi-seed stratified cross-validation procedure.

\subsubsection{Baseline Architectures and Parametric Matching}
\label{subsec:benchmark}

Three model classes are evaluated: a Kolmogorov-Arnold Network (KAN), a Multilayer Perceptrons (MLP), and an XGBoost ensemble. The KAN and MLP are compared under a strict parameter-parity constraint to ensure that performance differences reflect architectural properties rather than capacity differences. XGBoost is included as a robust, non-parametric baseline. Because gradient-boosted tree ensembles have no directly comparable parameter structure to neural networks, it is exempt from parameter matching, serving instead as a well-established discriminative reference point. All configurations are provided in Table~\ref{tab:phase2_architectures}.

\begin{table}[!htbp]
\centering
\caption{Model configurations and hyperparameter settings for the Phase II comparative benchmark}
\label{tab:phase2_architectures}
\scriptsize
\renewcommand{\arraystretch}{1.2}
\begin{tabular}{ll p{4.2cm}}
\toprule
\textbf{Model} & \textbf{Property} & \textbf{ Configuration} \\
\midrule
\textit{KAN}        & Architecture    & $[d_{\text{in}}, 10, D_{\text{out}} = 2]$ \\
                    & Spline settings & Cubic B-spline ($k = 3$), Grid resolution ($G_{\text{grid}} = 7$) \\
                    & Optimizer       & Adam ($\eta = 0.0005$, Weight decay $= 10^{-4}$) \\
\midrule
\textit{MLP}        & Hidden Width    & $W_{\text{MLP}}$ matched to KAN via \eqref{eq:mlp_single} \\
                    & Activation      & $\text{ReLU}$ \\
                    & Optimizer       & Adam ($\eta = 0.001$, Weight decay $= 10^{-5}$) \\
\midrule
\textit{XGBoost}
                    & Base Learners   & Gradient-boosted trees ($N_{\text{estimators}} = 100$) \\
                    & Learning rate   & $\eta = 0.1$ \\
                    & Loss hyperparameters  &  $\alpha_{\text{imb}}=\text{Class frequency weight}$, $\gamma =2.0$ \\
\midrule
\textit{Shared}     & Cross-Validation & Stratified 5-Fold \\
                    & Execution flags & Batch size $= 64$, epochs $= 100$ (maximum) \\
                    & Early stopping  & Patience $\mathcal{P} = 5$ epochs (KAN / MLP only) \\
                    & Random seeds    & $\mathcal{S} = \{42, 123, 999, 7, 21\}$ \\
\bottomrule
\end{tabular}
\end{table}

\paragraph{Kolmogorov-Arnold Network (KAN)}

The KAN architecture is strictly locked to the optimal configuration determined in Phase~I. Throughout training, a loss-based early stopping mechanism is applied to guarantee a  consistent convergence baseline. To isolate the effects of architectural behavior from initialization artifacts, performance is evaluated across a multi-seed stratified cross-validation protocol.

\paragraph{Analytical Multilayer Perceptron (MLP) }
The MLP serves as a primary parameter-matched  neural baseline. Its hidden width $W_{MLP}$ is recomputed at runtime for each dataset from the complied KAN parameter count ${N_{target}}$ via Equation~\eqref{eq:mlp_single}, maintaing parity as input dimensionality varies across the cohort. A single-layer depth is used throughout Phase~II, consistent with the KAN configuration, so that architectural depth is held constant and observed differences are attributed to the activation mechanism alone. The same optimizer settings and early stopping criterion are applied to both models.

\paragraph{Gradient Boosted Tree Ensemble (XGBoost)}

XGBoost is trained as a non-parametric reference under all three experimental conditions. Under the focal loss condition, a custom objective function is implemented via first-and second-order derivative (gradient and Hessian) injection into the boosting routine~\cite{kan46}. It provides XGBoost with a true focal loss objective consistent with the neural model implementation. The focal loss objective for training sample $i$ is:
\begin{equation}
\mathcal{L}_{\text{focal}}(\hat{y}_i, y_i) = -\alpha_{\text{imb}} y_i (1 - \hat{y}_i)^\gamma \log(\hat{y}_i) - (1 - y_i) \hat{y}_i^\gamma \log(1 - \hat{y}_i)
\end{equation}
where $y_i \in \{0, 1\}$ is the ground-truth binary label, and $\hat{y}_i$ is sigmoid transformed tree score. $\alpha_{imb}$ and $\gamma$ are the focusing parameter. Under the baseline and resampled conditions, the standard log-loss objective is used with no additional class weighting, since the resampled training distribution is already mitigated by SMOTE-Tomek. All other configurations are provided in Table~\ref{tab:phase2_architectures}.

\subsubsection{Imbalance Mitigation Strategies}
\label{subsec:interventions}

Phase II evaluates three training conditions across all models: a native baseline (no intervention), a data-level resampling condition, and a loss-level reweighting condition. The baseline establishes how each model responds to the native class distribution without any correction. Under this baseline, models are trained directly on the native distribution with no resampling, class weighting, or objective modification applied. This condition reveals the degree of majority-class bias each architecture exhibits without intervention. The two other active conditions then isolate whether KAN spline activations respond more effectively to distributional augmentation or to gradient-level reweighting.

\paragraph{SMOTE-Tomek resampling}
Under the resampled condition, a hybrid SMOTE-Tomek procedure is applied to the training fold of each cross-validation split prior to model fitting. SMOTE generates synthetic minority-class samples by interpolating between existing minority instances and their $\mathcal{k}$-nearest neighbors, increasing minority-class representation in the training set. Tomek Links undersampling then removes borderline cross-class pairs that lie within minimal mutual distance, reducing class overlap near the decision boundary. The combined effect increases minority-class density while cleaning ambiguous boundary regions. To prevent data leakage, resampling is applied strictly within each training fold and never to the held-out evaluation fold.

\paragraph{Focal Loss}
Under the loss-level condition, the standard cross-entropy objective is replaced with Focal Loss for all three models. It is applied as a batch-averaged loss for the KAN and MLP~\cite{kan48}:
\begin{equation}
    \mathcal{L}_{\text{focal}} = \frac{1}{\mathcal{B}} \sum_{i=1}^{\mathcal{B}} \left[ \alpha (1 - p_i)^{\gamma} \mathcal{L}_{\text{CE}}(\mathbf{z}_i, y_i) \right]
\end{equation}

where $p_i = \exp(-\mathcal{L}_{\text{CE}}(\mathbf{z_{i}, y_{i})} $ is the model's estimated probability for the true class of sample $i$, $\alpha$ is class frequency ratio, $\gamma = 2.0$ is the focusing exponent~\cite{kan39}. The XGBoost focal objective follows the same $\alpha$ and $\gamma$ conventions via modified gradient and Hessian statistics, ensuring consistency across model classes.

\subsubsection{Evaluation Metrics and Statistical Validation}
\label{subsec:statistical_validation}
All metrics are computed on the held-out validation fold only; with no resampling or class adjustment applied to the evaluation partition. Standard accuracy is unreliable under severe class imbalance since a classifier predicting only the majority class recall~\cite{kan24,kan27}. The framework therefore relies on three imbalance-sensitive metrics, summarized in Table~\ref{tab:METRIC}. The Geometric Mean ($G_{mean}$) serves as the primary evaluation criterion throughout, as it simultaneously penalizes poor performance on either class and collapse to zero if recall of any class approaches zero.

\begin{table}[!htbp]
\centering
\caption{Evaluation metrics and their sensitivity to class imbalance.}
\label{tab:METRIC}
\scriptsize 
\renewcommand{\arraystretch}{1.3}
\begin{tabularx}{\columnwidth}{llcX}
\toprule
\textbf{Metric} & \textbf{Sensitivity} & \textbf{Range} & \textbf{Role} \\
\midrule
$G_{\text{mean}}$ & Very high & $[0,1]$ & Primary; penalizes collapse of minority-class recall \\
Macro $F_1$ & High & $[0,1]$ & Joint precision--recall reliability across both classes \\

AUC-ROC & Moderate & $[0.5,1]$ & Threshold-independent rank discrimination \\
\bottomrule
\end{tabularx}
\end{table}


\paragraph{Statistical Validation}
As the benchmark spans ten structurally distinct datasets, standard parametric tests are not appropriate since normality of performance distributions cannot be assumed across datasets with such different characteristics~\cite{kan50}. A two-stage non-parametric inference procedure is therefore applied, treating the datasets as independent sampling units. 

First, a Friedman test~\cite{kan51} evaluates the global null hypothesis that all model-strategy configurations perform equivalently across the cohort. Let $n=10$ denote the number of evaluation datasets, and $R_{j}$ the cumulative rank sum for the $j^{th}$ configuration. The test statistic is:

\begin{equation}
    \chi_F^2 = \frac{12}{nk(k+1)} \sum_{j=1}^{k} R_j^2 - 3n(k+1)
\end{equation}

Upon rejection of the global null hypothesis ($\alpha = 0.05$), targeted pairwise comparisons are conducted using the two-tailed Wilcoxon signed-rank test~\cite{kan51}. To quantify the practical magnitude of observed differences independently of sample size, each comparison is accompanied by Cohen's $d_z$ effect size~\cite{kan52}, computed over paired differences across datasets:
\begin{equation}
d_z = \frac{\bar{\delta}}{s_\delta}
\end{equation}
where $\bar{\delta}$ is the mean of the paired differences ($x_{i}-y_{i}$) across datasets and ${s_\delta}$ is their standard deviation. By convention, the first-listed configuration is used as the minuend ($x_{i}$). Consequently, for higher-is-better metrics (e.g.,$G_{mean}$), a negative $d_z$ indicates improvement brought by the second configuration over the reference. Conversely, for lower-is-better metrics (such as training latency) a positive $d_z$ indicates that the second configuration reduces the metric value, thereby representing a gain over the reference. Effect size magnitudes are categorized by standard absolute thresholds: negligible ($|d_z|<0.2$), small ($0.2\leq|d_z|<0.5$), medium ($0.5\leq |d_z|<0.8$), or large $|d_z|>0.8$. 

To provide a comprehensive overview of global performance across all conditions, we compute the macro-averaged mean rank $\mathcal{R}$ for each model strategy configuration across the ten-dataset cohort. Here, a lower value indicates better overall performance. These relative hierarchies are reported alongside pairwise validation result. 

\section{Experimental Results and Analysis}
\label{sec:results}

The section presents the empirical findings from the two-phase evaluation framework. Section~\ref{subsec:phase1_results} reports the Phase I spline grid sensitivity analysis, which establishes the spline configuration for Phase II. Section~\ref{subsec:phase2_results} reports the Phase II comparative benchmark across the ten datasets.

\subsection{Phase I: Spline Grid Sensitivity Analysis}
\label{subsec:phase1_results}

\begin{figure*}[!htbp]
\centering

\includegraphics[width=\textwidth]{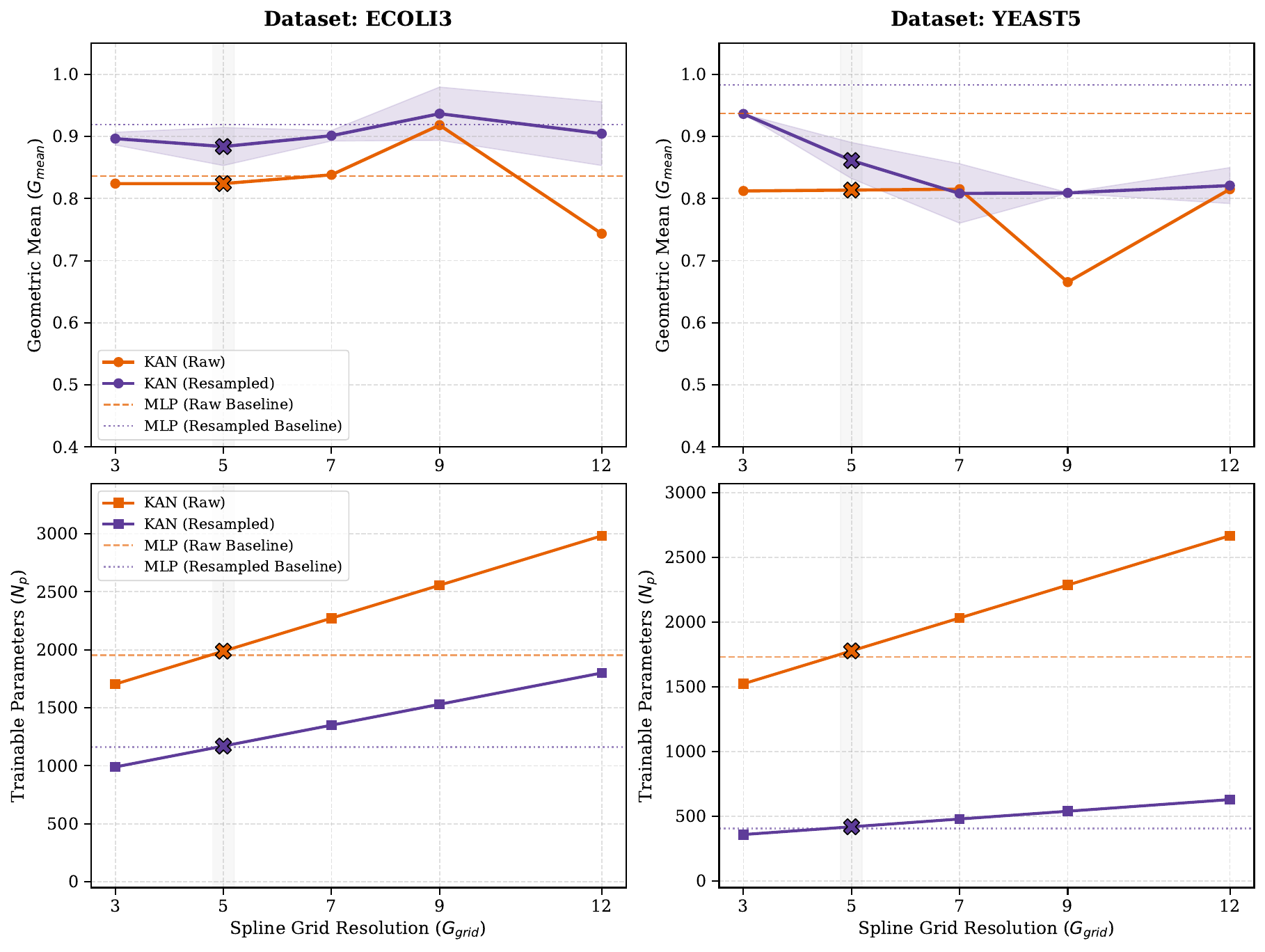}
\caption{Validation $G_{\text{mean}}$ (upper panels) and trainable parameter count $N_p$ (lower panels) as a function of spline grid resolution $G_{\text{grid}} \in \{3, 5, 7, 9, 12\}$ for \texttt{ecoli3} and \texttt{yeast5}, under baseline (Raw) and SMOTE-Tomek resampled training conditions. All hyperparameters other than $G_{\text{grid}}$ are frozen to HPO-selected values. Shaded bands denote cross-seed variance. Cross markers $(\times)$ indicate the HPO anchor at $G_{\text{grid}}=5$; horizontal dashed and dotted lines show the fixed MLP parameter budget and performance at parity.}
\label{fig:phase1_curves}
\end{figure*}

The Phase~I sweep evaluates KAN classification performance ($G_{\text{mean}}$) and trainable parameter count ($N_p$) across grid resolutions $G_{\text{grid}} \in \{3, 5,7,9,12\}$ with all other hyperparameters frozen to the values identified by the HPO procedures at anchor $G_{\text{grid}}=5$. Fig.~\ref{fig:phase1_curves} shows the multi-seed performance profiles (upper panels) and parameter scaling trajectories (lower panels) for both calibration datasets under baseline and SMOTE-Tomek resampled training conditions. The flat horizontal reference lines show the fixed MLP parameter budget and performance at the parity anchor, providing a stable comparison baseline as the KAN grid resolution changes. 

The trajectories in Fig.~\ref{fig:phase1_curves} reveal consistent patterns across both datasets that motivate the grid resolution selected for Phase~II.

\paragraph{Parametric Count Scaling and Architectural Compactness}

The parameter count ($N_p$) increases linearly with $G_{\text{grid}}$ across all evaluated resolutions. Critically, it directly reflects the expansion of the underlying B-spline coefficient space per network edge. The HPO pipeline independently selected structurally distinct hidden layer configurations for the two training conditions. As a result, the resampled and baseline models trace completely distinct, non-overlapping parameter allocation paths throughout the resolution spectrum. This clear separation shows that the single-layer configurations optimized for resampled feature spaces maintain a lower capacity regime than their baseline counterparts. Nevertheless, both scale predictably with grid refinement.

For \texttt{ecoli3}, the baseline condition (a two-layer network with widths $[10, 6]$) yileds a KAN parameter count that rises from approximately $1{,}700$ at $G_{\text{grid}}=3$ to $3{,}000$ at $G_{\text{grid}}=12$, consistently above the fixed MLP reference of $1{,}952$. The resampled condition, where HPO converged to a shallower single-layer network of width 10, occupies a more compact range of approximately $1{,}000$ to $1{,}800$ parameters across the same spectrum, closely tracking the fixed MLP resampled reference of $1{,}162$.

For \texttt{yeast5}, the baseline architecture (two layers, widths $[9, 5]$) scales from approximately $1{,}550$ to $2{,}650$ parameters, again exceeding the MLP reference of $1{,}730$. The resampled condition, where HPO selected an exceptionally compact single-layer network of width~3, produces a substantially smaller range of approximately $360$ to $625$ parameters (well below the MLP resampled reference of $407$) reflecting the minimal architectural footprint that suffices under augmented minority coverage.

This architectural divergence between conditions is itself a substantive observation. It shows that SMOTE-Tomek's distributional augmentation alters not only training outcomes but the capacity requirements of the spline network. Under the baseline condition, the sparse minority signal forces HPO toward wider, deeper networks. This expansion allocates more edge functions to the minority region in hopes of encountering sufficient activation mass. Conversely, the denser minority coverage under the resampled condition allows a compact network with fewer total edges to receive adequate minority-class activation at each spline knot. This rich coverage allows individual edge functions to develop non-trivial curvature without requiring additional layers to compensate.

\paragraph{Performance Trajectories and Variance Structure}
The $G_{\text{mean}}$ profiles in Fig.~\ref{fig:phase1_curves} reveal qualitatively different stability behaviors. Across both datasets, $G_{\text{grid}}=7$ emerges as the critical resolution where performance stabilizes and remains consistent. 

For \texttt{ecoli3}, the baseline KAN shows a gradual rise from $G_{\text{mean}} \approx 0.82$ at $G_{\text{grid}}=3$ through $0.84$ at $G=7$, then a sharp peak at $G=9$ (${\approx}0.92$) followed by an equally sharp collapse to $0.75$ at $G=12$. This non-monotonic trajectory is a characteristic signature of spline overfitting. At $G=9$, the fine grid resolution allows the spline activations to closely fit the training fold's class boundaries, yielding inflated validation performance on specific seeds. However, this localized memorization fails to generalize,  causing the sharp collapse at $G=12$ as the knot density exceeds what the 336-sample dataset can stably support. The resampled KAN for \texttt{ecoli3} is substantially more stable, maintaining $G_{\text{mean}}$ between $0.88$ and $0.95$ across the full spectrum with narrow cross-seed variance. This stability confirms that the augmented minority distribution suppresses the overfitting instability that the baseline condition exposes.

For \texttt{yeast5}, the dominant feature is a severe trough in the baseline KAN at $G=9$, where $G_{\text{mean}}$ drops to approximately $0.67$ (more than 0.15 points below its value at adjacent resolutions). Under severe imbalance ($\text{IR}=32.73$), the native training distribution provides minimal minority-class activation for the spline edge functions. At higher grid resolutions, this sparsity is compounded. Additional knot intervals receive even fewer minority-class observations, causing the corresponding spline coefficients default to uninformative fits. The result is an activation landscape in which the majority boundary is well characterized but the minority region collapses unpredictably depending on which seeds happen to place training samples near the relevant knots. 

The resampled condition for \texttt{yeast5} shows a different pattern: $G_{\text{mean}}$ begins relatively high at $G=3$ (${\approx}0.94$) but with the widest variance band visible in the figure. It then declines and stabilizes from $G_{\text{grid}}=7$ onward at approximately $0.81$ with markedly tighter cross seed variance. The high mean at $G=3$ is therefore unreliable, as the coarse grid makes performance sensitive to initialization with some seeds achieving high recall and others failing. In contrast, the plateau from $G=7$ onward represents genuinely stable minority modeling.

Across both datasets, the MLP reference lines remain flat by construction, as the MLP architecture is fixed independently of $G_{\text{grid}}$. These flat references serve two functions. First, they confirm that the parity anchor is maintained as the KAN resolution varies. Second, they provide a performance benchmark against which the KAN trajectories can be read. At $G_{\text{grid}}=7$, the KAN resampled condition meets or closely approaches the MLP resampled reference on both datasets without the parameter inflation seen at higher resolutions.

\paragraph{Grid Resolution Selection and Transfer to Phase~II}

The two behavioral patterns identified above jointly eliminate the extreme resolution values as viable candidates. These include the overfitting instability of the baseline KAN at $G \geq 9$ for \texttt{ecoli3}, and the variance-driven unreliability at $G=3$ for \texttt{yeast5} resampled. Between the remaining options, $G_{\text{grid}}=7$ is the first resolution at which both datasets simultaneously exhibit stable performance and narrow cross-seed variance under the resampled condition. Crucially, this choice achieves stability without exceeding the parameter budget that delivers diminishing returns in the baseline condition. Selecting $G=9$ would recover the ecoli3 baseline peak but at the cost of the yeast5 trough ($G_{\text{mean}} \approx 0.67$). Conversely, selecting $G=5$ leaves measurable variance in the yeast5 resampled band that has not yet converged to the tight range seen at $G=7$. $G_{\text{grid}}=7$ is therefore the resolution at which cross-dataset consistency and parameter efficiency jointly stabilize.

The grid sweep was conducted under SMOTE-Tomek conditions, which provide the densest minority coverage. Consequently, the resolution at which saturation occurs ($G=7$) transfers conservatively to the sparser baseline and Focal Loss conditions in Phase~II. Under these sparser conditions, where minority coverage is lower, the selected anchor grid resolution provides at least adequate knot density without risk of under-resolution.

\subsection{Phase II: Comparative Benchmark Results}
\label{subsec:phase2_results}

Phase~I established that $G_{\text{grid}}=7$ is the resolution at which KAN spline activations reach stable minority-class coverage under resampled conditions. Furthermore, it demonstrated that shallower architectures suffice when the training distribution is augmented. Phase~II now applies that fixed configuration across the full ten-dataset cohort to answer the central question. How do KAN spline activations interact with data-level and loss-level imbalance interventions relative to parameter-matched and ensemble baselines, and at what computational cost?

\subsubsection{Classification Performance Across Models and Conditions}
\label{subsubsec:perf_space}

\begin{figure}[!htbp]
\centering
\includegraphics[width=\columnwidth]{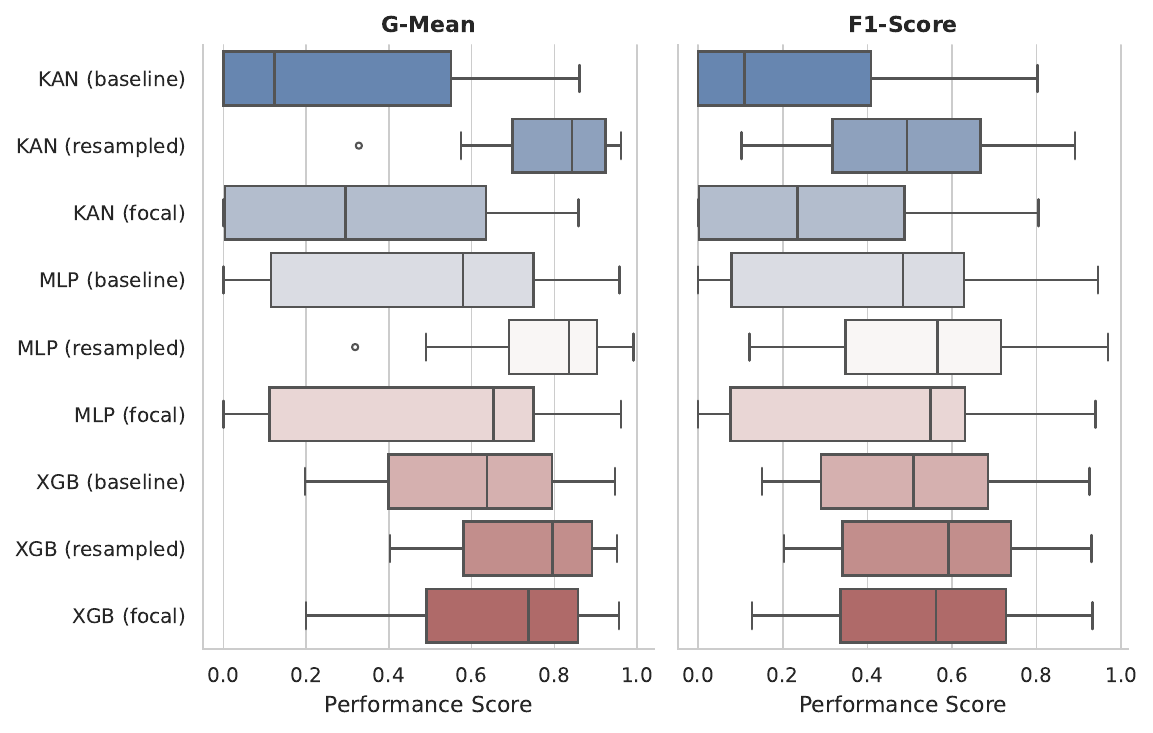}
\caption{Per-dataset distributions of $G_{\text{mean}}$ (left) and Macro $F_1$-Score (right) across all model--strategy configurations, showing median, interquartile range, and individual dataset observations. The vertical spread within each configuration reflects performance variation across the imbalance spectrum of the cohort ($\text{IR} = 5.14$--$46.50$).}
\label{fig:ieee_metrics}
\end{figure}

Macro-averaged $G_{\text{mean}}$, Macro~$F_1$, and AUC-ROC across the ten-dataset cohort are reported in Table~\ref{tab:global_performance}, alongside the minimum per-dataset $G_{\text{mean}}$ as a fragility indicator. Per-dataset distribution spreads are shown in Fig.~\ref{fig:ieee_metrics}.

\begin{table}[!htbp]
\centering
\caption{Macro-averaged classification performance and minimum $G_{\text{mean}}$ across the ten-dataset cohort}
\label{tab:global_performance}
\footnotesize 
\setlength{\tabcolsep}{3.1pt} 
\renewcommand{\arraystretch}{1.2}
\begin{tabularx}{\columnwidth}{llcccc}
\toprule
\makecell[b]{\textbf{Model}} & \makecell[b]{\textbf{Strategy}} & \makecell[c]{\textbf{G-Mean}\\\textbf{(\%)}} & \makecell[b]{\textbf{Macro $F_1$}\\\textbf{(\%)}} & \makecell[b]{\textbf{AUC-ROC}\\\textbf{(\%)}} & \makecell[b]{\textbf{Min.}\\\textbf{$G_{\text{mean}}$}} \\
\cmidrule(lr){3-3}
& & \textbf{Mean $\pm$ Std.} & & & \\
\midrule
KAN        & baseline   & 28.64 $\pm$ 34.80 & 24.31 $\pm$ 30.91 & 85.75 $\pm$ 13.05 & 0.00 \\
           & focal      & 34.24 $\pm$ 34.49 & 29.13 $\pm$ 30.89 & 87.52 $\pm$ 11.47 & 0.00 \\
           & resampled  & 78.60 $\pm$ 18.62 & 50.60 $\pm$ 26.17 & 90.45 $\pm$ 8.97 & 39.78 \\
\hline
MLP        & baseline   & 49.48 $\pm$ 37.65 & 43.14 $\pm$ 35.46 & 88.84 $\pm$ 9.07 & 0.00 \\
           & focal      & 51.02 $\pm$ 38.27 & 44.23 $\pm$ 35.76 & 89.45 $\pm$ 8.38 & 0.00 \\
           & resampled  & 76.44 $\pm$ 20.43 & 54.42 $\pm$ 28.23 & 89.09 $\pm$ 8.70 & 37.50 \\
\hline
XGB        & baseline   & 60.10 $\pm$ 24.86 & 50.83 $\pm$ 25.75 & 91.16 $\pm$ 6.89 & 22.36 \\
           & focal      & 66.53 $\pm$ 23.91 & 53.53 $\pm$ 25.52 & 91.24 $\pm$ 6.55 & 22.28 \\
           & resampled  & 74.51 $\pm$ 17.57 & 56.95 $\pm$ 24.05 & 92.17 $\pm$ 6.12 & 46.43 \\
\bottomrule
\end{tabularx}
\end{table}

The results reveal distinct performance patterns across model classes and training conditions:

\paragraph{Baseline Condition}

Without any imbalance mitigation, the KAN exhibits the most severe majority-class bias of the three models. Its macro-averaged $G_{\text{mean}}$ of $28.64\% \pm 34.80\%$ conceals a binary pattern visible in Fig.~\ref{fig:ieee_metrics}. On five of the ten datasets the KAN fails entirely ($G_{\text{mean}} = 0.00\%$), all of which carry IR $\geq 8.6$ (\texttt{ecoli3}, \texttt{glass2}, \texttt{yeast6}, \texttt{winequality-red-8\_vs\_6}, \texttt{winequality-red-8\_vs\_6-7}).

On the remaining five (with some lower-IR datasets) it achieves
competitive recall. This bimodal distribution, rather than uniformly poor performance, is structurally informative. The B-spline activations develop meaningful curvature on datasets where minority density is sufficient for local coefficient updates, but collapse to flat majority-dominated responses where
it is not. Crucially, AUC-ROC remains elevated on these failed datasets (e.g., $0.942$ on \texttt{yeast6}, $0.886$ on \texttt{ecoli3}). This discrepancy confirms that the model retains probabilistic class separation while its decision threshold is overwhelmed by the majority class. The parameter-matched MLP baseline ($49.48\% \pm 37.65\%$) also reaches $G_{\text{mean}} = 0.00\%$ on three datasets under extreme IR. However, its fixed ReLU activations and globally coupled weight matrices exhibit less pronounced sensitivity to minority-class sparsity than the locally adaptive spline functions, yielding a higher cohort average despite sharing the same parameter budget.

\paragraph{Focal Loss Condition}
Focal Loss produces selective rather than uniform improvement across the cohort. The KAN improves from $28.64\%$ to $34.24\%$ overall, but this aggregate masks where the gain concentrates. On \texttt{yeast6} (IR $= 41.40$) the KAN recovers from $0.00\%$ to $33.37\%$, and on \texttt{yeast4} (IR $= 28.10$) it rises by $10.5$ percentage points. On \texttt{winequality-red-8\_vs\_6-7} (IR $= 46.50$) and \texttt{winequality-red-8\_vs\_6} (IR $= 35.44$), the KAN remains at $0.00\%$ under focal loss. These represent the most extreme imbalance cases in the cohort where the spline activations still encounter insufficient minority mass regardless of gradient reweighting. This pattern reveals the structural limit of loss-level intervention for KANs. While focal loss can redirect gradient pressure toward minority samples that are present but misclassified, but it cannot compensate for the near-absence of minority observations in the activation domain of individual spline edges. The MLP shows a similar but attenuated pattern, improving marginally to $51.02\%$ while retaining $0.00\%$ on the same extreme-IR datasets.

\paragraph{Resampled Condition}
SMOTE-Tomek resampling produces the most substantial and qualitatively different shift in the benchmark. The resampled KAN achieves $78.60\% \pm 18.62\%$, eliminating all zero-$G_{\text{mean}}$ failures across the cohort (minimum $39.78\%$), and outperforming both the parameter-matched MLP ($76.44\%$) and XGBoost ($74.51\%$). The performance reversal is striking. Here, the KAN, which ranks last under the baseline condition, ranks first under resampling. The mechanism is consistent with the Phase~I finding that spline activations require a minimum density of minority-class observations to develop non-trivial curvature. By augmenting the training distribution rather than only reweighting the loss, SMOTE-Tomek provides that density. It enables each edge's B-spline function to fit the minority-class decision region with precision unavailable to fixed-basis MLPs. However, this advantage is not uniform. On low-IR datasets (\texttt{new-thyroid1} IR $= 5.14$, \texttt{ecoli3} IR $= 8.60$), the MLP matches or slightly outperforms the KAN, suggesting that fixed activations suffice when the minority-class signal is already moderate. The KAN's advantage is not uniform, however. The MLP matches or exceeds the KAN on the three lowest-IR datasets in the cohort: \texttt{new-thyroid1}($\text{IR}=5.14$), \texttt{ecoli3} ($\text{IR}=8.60$), and \texttt{glass2}($\text{IR}=11.59$). On the remaining seven datasets the KAN leads by margins ranging from $0.5$ to $10.4$ percentage points, with the largest gains
concentrated on \texttt{yeast4} ($+10.3$~pp) and \texttt{winequality-red-8\_vs\_6} ($+10.4$~pp), precisely the high-IR datasets where distributional augmentation most expands the minority activation domain of individual spline edges.



\subsubsection{Multi-Metric Profile- AUC--$G_{\text{mean}}$ Decoupling}
\label{subsubsec:intervention_analysis}

\begin{figure}[!htbp]
\centering
\includegraphics[width=\columnwidth]{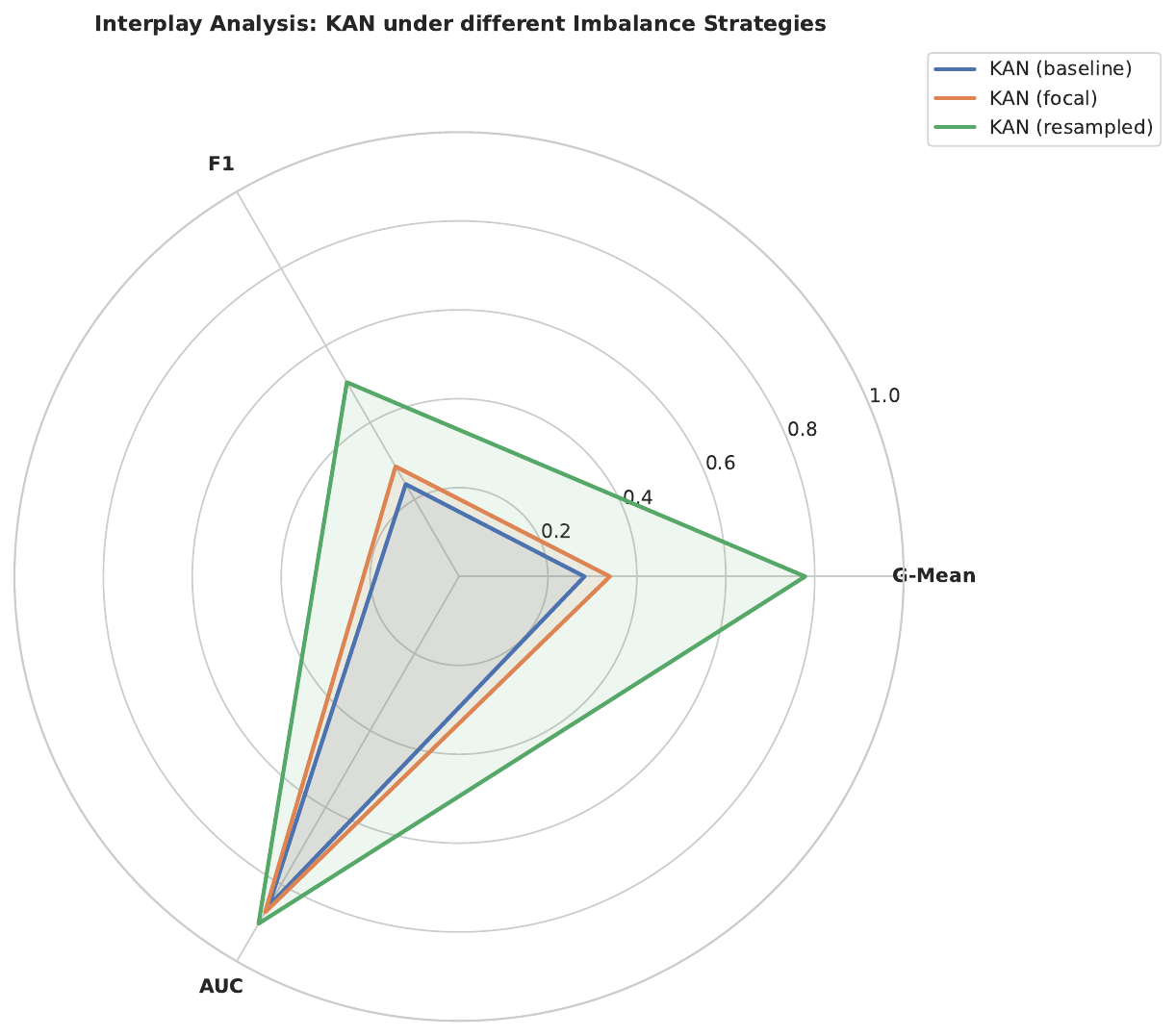}
\caption{Macro-averaged multi-metric performance profiles of the KAN under baseline, focal, and SMOTE-Tomek resampled conditions, showing $G_{\text{mean}}$, Macro $F_1$-Score, and AUC-ROC simultaneously. The area of each profile reflects the degree of alignment between threshold-independent rank separation and threshold-dependent minority-class classification.}
\label{fig:kan_radar}
\end{figure}

The three-metric radar plot in Fig.~\ref{fig:kan_radar} reveals a pattern not visible from table values alone. The relationship between threshold-free ranking quality (AUC-ROC) and threshold-dependent minority recall ($G_{\text{mean}}$, Macro~$F_1$) changes structurally, not just quantitatively, across conditions.

Under the baseline condition, the radar profile is highly asymmetric. AUC-ROC of $85.75\%$ is preserved while $G_{\text{mean}}$ and Macro~$F_1$ collapse to $28.64\%$ and $24.31\%$ respectively. This decoupling (a gap of over 57 percentage points between AUC-ROC and $G_{\text{mean}}$) reflects a specific failure mode. The model has learned to rank minority instances with moderate confidence in its probability outputs, but the decision boundary is positioned entirely within the majority class. The AUC remains high because it is threshold-independent. Meanwhile, the classification output fails because the threshold operates in a region where no minority-class discrimination occurs. This is not a generic imbalance failure; it is specific to the spline mechanism. As each edge function is fitted locally, activations in minority-sparse regions remain flat and contribute negligible gradient signal at the boundary, while the model's ranking behavior is preserved in the probabilistic output layer.

The focal loss profile expands slightly and uniformly, indicating a consistent shift in the decision threshold rather than a change in the underlying representation. The triangular shape remains similar to baseline, confirming that the loss-level intervention does not alter the fundamental AUC--$G_{\text{mean}}$ structure. It moves the threshold but does not resolve the activation collapse in minority-sparse spline regions.

The resampled profile is qualitatively distinct. A broader, more balanced triangle in which all three axes improve simultaneously, with the AUC--$G_{\text{mean}}$ gap narrowing from 57 to 12 percentage points ($90.45\%$ vs $78.60\%$). This simultaneous convergence of threshold-free and threshold-dependent performance is the signature of improved underlying class separation, not threshold recalibration. It confirms that SMOTE Tomek does not merely shift the operating point, it also enables the spline activations to develop genuine minority-class boundary structure that persists regardless of threshold placement.


\subsubsection{Computational Cost Analysis}
\label{subsubsec:hardware_efficiency}

\begin{figure}[!htbp]
\centering
\includegraphics[width=\columnwidth]{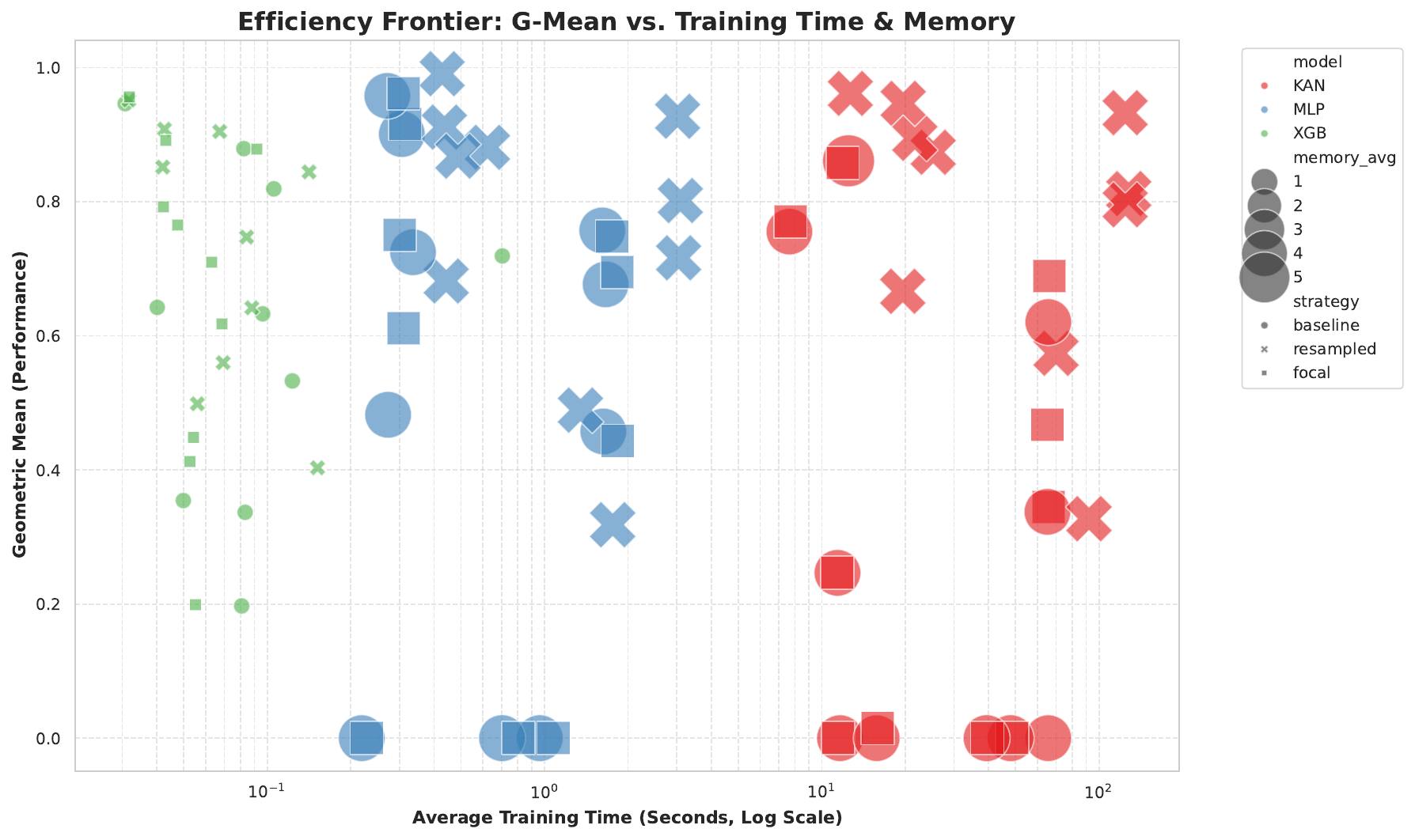}
\caption{$G_{\text{mean}}$ vs.\ average training time (log scale) for all model--strategy configurations. Bubble size encodes peak memory usage. All values are aggregated across the ten-dataset cohort.}
\label{fig:resource_frontier}
\end{figure}

Fig.~\ref{fig:resource_frontier} maps $G_{\text{mean}}$ against average
training time (log scale) and memory footprint across all configurations. This visualization reveals that the three model classes occupy structurally separate regions of
the performance--cost space.

\paragraph{XGBoost}
XGBoost occupies the left-hand extreme of the frontier, training in $0.03$--$0.70$ seconds across all conditions with minimal memory overhead. Its macro averaged $G_{\text{mean}}$ spans from $60.10\%$ (baseline) to $74.51\%$ (resampled) providing a competitive performance at negligible computational cost. The absence of backpropagation and the O(nlog n) complexity of tree-split search accounts for this efficiency. Notably, XGBoost achieves this competitive performance without ever failing completely across the cohort. As, its minimum per-dataset $G_{\text{mean}}$ under the baseline condition is 22.36\%, in contrast to the complete classification failures observed for both neural architectures on the most severely imbalanced datasets.

\paragraph{MLP}
The MLP cluster spans $0.22$--$3.09$ seconds, with memory stable across all three conditions because weight matrix size is fixed regardless of training strategy. The cluster shows substantial vertical spread in Fig.~\ref{fig:resource_frontier}. The resampled configuration reaches $76.44\%$ $G_{\text{mean}}$ while the baseline and focal configurations collapse to near-zero on the most severely imbalanced datasets, all within the same training time envelope. This confirms that the MLP's resource footprint is predictable but does not protect against performance fragility under extreme imbalance.

\paragraph{KAN}
The KAN configurations occupy the right-hand region, with training times of $7.7$--$130.7$ seconds per fold, a factor of $30$--$43\times$ slower than the parameter-matched MLP across all datasets and conditions. This overhead is consistent across strategies (mean ratio: $42.8\times$ baseline, $39.0\times$ focal, $43.8\times$ resampled) and arises from the spline evaluation procedure. Each forward pass requires computing B-spline basis functions at every grid point across every edge, an operation with no equivalent fused CUDA kernel in the current \texttt{pykan} implementation. This stands in sharp contrast to the matrix multiplications that underlie MLP inference. Under the resampled condition, KAN training times extend to $70$--$128$ seconds on the larger yeast and winequality datasets, reflecting the increased epoch cost of the augmented training set. The resampled KAN configurations occupy the upper boundary of the frontier in $G_{\text{mean}}$. However, their bubble sizes are also the largest, confirming that peak memory usage scales with training set size under SMOTE-Tomek augmentation. The efficiency frontier therefore does not yield a Pareto-dominant solution. The resampled KAN achieves the highest classification performance but at a cost that is one to two orders of magnitude above both baselines.

\subsubsection{Statistical Validation Results}
\label{subsubsec:statistical_validation_results}


\paragraph{Global Significance}
The Friedman test across all $k=9$ model--strategy configurations and $n=10$ datasets yields $\chi_F^2 = 63.68$ ($p < 0.0001$). This result rejects the null hypothesis of equivalent performance, confirming that the observed ranking differences are not attributable to stochastic variability. Mean ranks $\mathcal{R}$ across the cohort are reported in Table~\ref{tab:mean_ranks}.

\begin{table}[!htbp]
\centering
\caption{Mean performance ranks across the ten-dataset cohort (lower is better)}
\label{tab:mean_ranks}
\scriptsize
\renewcommand{\arraystretch}{1.2}
\begin{tabular}{llc}
\toprule
\textbf{Model} & \textbf{Strategy} & {$\mathcal{R}$} \\
\midrule
KAN   & Resampled & 1.70 \\
MLP   & Resampled & 2.00 \\
XGB   & Resampled & 3.10 \\
XGB   & Focal     & 4.30 \\
XGB   & Baseline  & 5.80 \\
MLP   & Focal     & 5.85 \\
MLP   & Baseline  & 6.35 \\
KAN   & Focal     & 7.80 \\
KAN   & Baseline  & 8.40 \\
\bottomrule
\end{tabular}
\end{table}

The rank table reveals two structural patterns. First, the top three ranks are occupied exclusively by resampled configurations (regardless of model class) confirming that data-level distributional augmentation is the dominant factor in cohort-wide performance, not architectural choice. Second, within each strategy tier, the ordering inverts across conditions. KAN ranks last among baseline configurations ($\mathcal{R}=8.40$) but first among resampled ones ($\mathcal{R}=1.70$). This swing of seven rank positions is the largest condition-induced rank shift of any model. XGBoost shows the most stable ranking across conditions ($\mathcal{R}=5.80$--$3.10$), consistent with its tree-ensemble structure's inherent resistance to distributional shift.

\paragraph{Pairwise Comparisons}

Targeted pairwise Wilcoxon signed-rank tests with Cohen's $d$ effect sizes are reported in Table~\ref{tab:statistical_outcomes}. Three findings are statistically and practically significant.

\begin{table}[!htbp]
\centering
\caption{Pairwise Wilcoxon signed-rank test results and Cohen's $d$ effect sizes ($\mu_{\text{Condition 1}} - \mu_{\text{Condition 2}}$) across the ten-dataset cohort.}

\label{tab:statistical_outcomes}
\scriptsize
\setlength{\tabcolsep}{3pt}
\renewcommand{\arraystretch}{1.3} 
\begin{tabularx}{\columnwidth}{llccc}
\toprule
\makecell[b]{\textbf{Metric}} & 
\makecell[b]{\textbf{Comparison}} & 
\makecell[b]{\textbf{$p$-value}} & 
\makecell[b]{\textbf{Cohen's $d$}} & 
\makecell[b]{\textbf{Outcome}} \\
\midrule
$G_{\text{mean}}$ & KAN Baseline vs.\ KAN Focal     & 0.0312 & $-0.17$ & Neg.-Sig. \\
\hline
$G_{\text{mean}}$ & KAN Baseline vs.\ KAN Resampled & 0.0020 & $-1.75$ & Sig. \\
\hline
$G_{\text{mean}}$ & KAN Baseline vs.\ MLP Baseline  & 0.0180 & $-0.59$ & Sig. \\
\hline
$G_{\text{mean}}$ & KAN Resampled vs.\ MLP Resampled & 0.2754 & $+0.10$ & \makecell[l]{Non-sig.} \\
\hline
$G_{\text{mean}}$ & KAN Resampled vs.\ XGB Resampled & 0.0645 & $+0.23$ & Non-sig. \\
\hline
\makecell[l]{Latency}          & KAN Baseline vs.\ MLP Baseline  & 0.0020 & $\hphantom{}+1.88$ & Sig. \\
\hline
Latency  & KAN Resampled vs.\ MLP Resampled  & 0.0020 & $+1.78$ & Sig. \\
\bottomrule
\end{tabularx}
\end{table}

\begin{enumerate}[label=(\roman*), leftmargin=*, noitemsep, topsep=2pt]
    \item \textit{Resampling effect on KAN:} The KAN baseline-to-resampled shift is both statistically significant ($p=0.0020$) and large in effect ($d=-1.75$). In contrast, the baseline-to-focal shift is significant at the $\alpha=0.05$ threshold ($p=0.0312$) but yields a negligible effect size ($d=-0.17$). This disparity confirms that the $5.6$ percentage-point mean improvement from focal loss is not practically consistent across the cohort. This shapes quantitative distinction in the benchmark as data-level resampling produces a large, reproducible gain; loss level reweighting produces a marginal one.
    
    \item \textit{KAN vs. MLP under the baseline condition:}
    The MLP baseline significantly outperforms the KAN baseline ($p=0.0180$, $d=-0.59$, medium effect). This confirms that without distributional correction, the globally-coupled weight matrices of the MLP are more robust to majority-class bias. Consequently, the locally adaptive spline functions, which require per-edge minority coverage to update reliably.
   
    \item \textit{Performance parity and Computational Cost under resampling:}
    Under the resampled condition, the KAN and MLP show no statistically significant $G_{\text{mean}}$ difference ($p=0.2754$, $d=+0.10$), meaning the macro-averaged advantage of the KAN does not constitute a globally consistent lead across all ten datasets. Yet the latency comparison under both baseline ($d=+1.88$) and resampled ($d=+1.78$) conditions is significant and large, confirming that the training overhead is a stable, architecture-level cost independent of training strategy. The resampled condition does not widen the latency gap further in effect-size terms, despite the larger training set, because the MLP's training time also scales with augmented data volume. Similarly, the comparison between the resampled KAN and resampled XGBoost ($p=0.0645$, $d=+0.23$, non-significant) confirms that the KAN's macro-averaged lead over XGBoost under resampling is also not cohort-consistent. The resampled KAN achieves the best mean rank ($\mathcal{R}=1.70$) without demonstrating a statistically uniform advantage over either baseline.

\end{enumerate}

\section{Discussion}
\label{sec:discussion}

The two-phase evaluation establishes that KAN performance under class imbalance is strongly condition-dependent in a structured and mechanistically explicable way. Under native class distributions, KAN spline activations are more susceptible to majority-class collapse than a parameter-matched MLP. Under
SMOTE-Tomek resampling, they achieve the best cohort-wide rank at statistically significant and large effect size. This section interprets the internal mechanism behind that reversal through direct inspection of the learned activation functions, then synthesizes the performance and cost findings into a deployment characterization.

\subsection{B-Spline Activation Analysis}
\label{subsec:manifold_audit}

\begin{figure}[!htbp]
\centering
\includegraphics[width=\columnwidth]{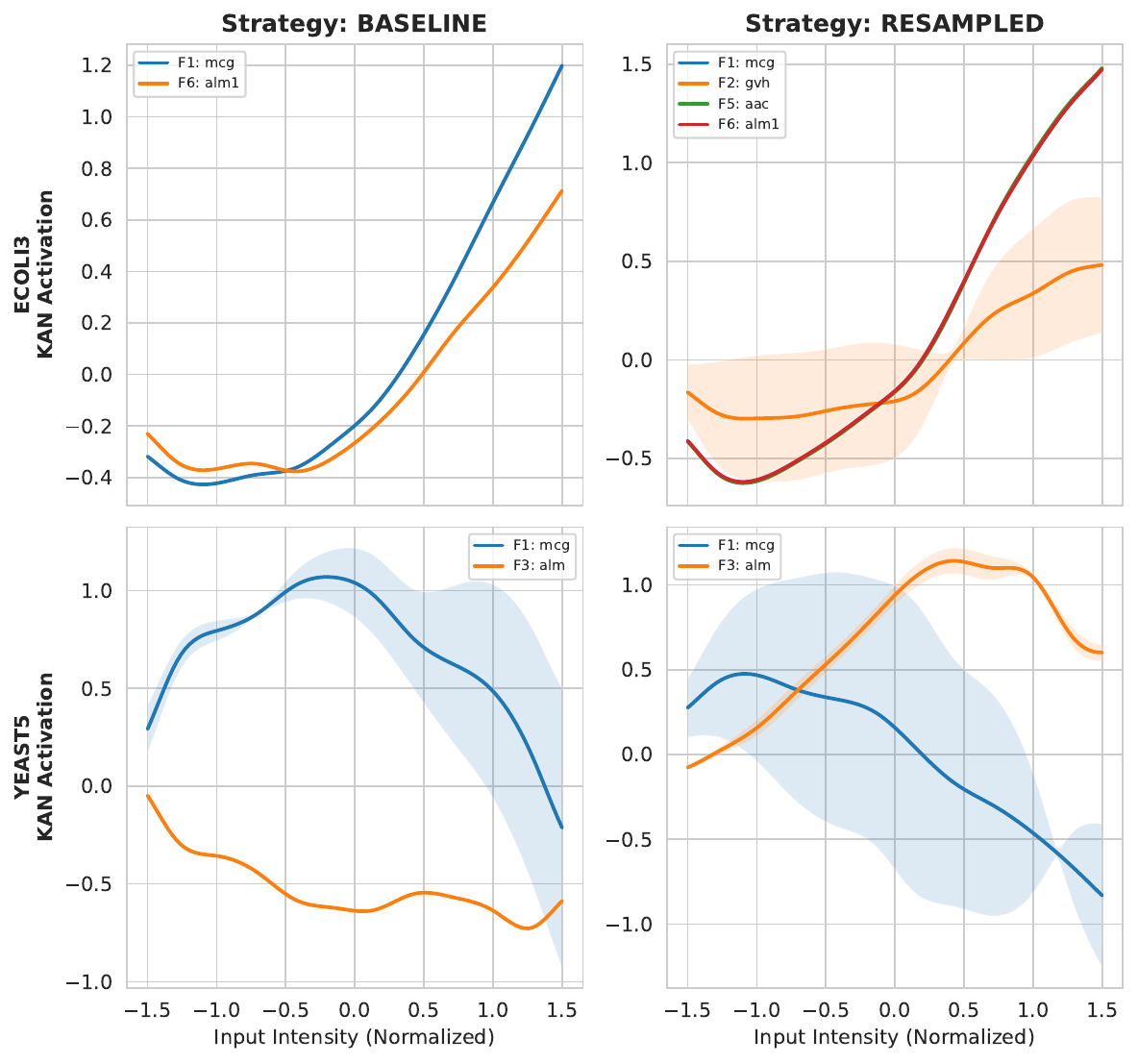}
\caption{Learned B-spline activation functions $\phi_{i,j}(x)$ across hidden-layer edges for \texttt{ecoli3} (top) and \texttt{yeast5} (bottom), under baseline (left) and SMOTE-Tomek resampled (right) training conditions. Shaded bands indicate cross-seed standard deviation. Only edges whose activation range exceeds a minimum threshold are shown; suppressed edges contribute negligible  majority class dominated optimization.}
\label{fig:spline_audit}
\end{figure}

KANs parameterize each network connection as a learnable univariate B-spline $\phi_{i,j}(x)$, meaning the activation function itself (not a weight applied to a fixed function) is what the model optimizes. This structure is the architectural basis for the interpretability claim associated with KANs. Each edge function can be read directly as the model's learned transformation of a single input feature, without requiring post-hoc attribution. To make this concrete and to connect the Phase~II performance results to the internal behavior of the model, Fig.~\ref{fig:spline_audit} shows the learned $\phi_{i,j}(x)$ profiles across the most active hidden-layer edges. These profiles are evaluated for \texttt{ecoli3} ($\text{IR}=8.60$) and \texttt{yeast5} ($\text{IR}=32.73$) under baseline and resampled training conditions. These are the same two datasets used in Phase~I. They were selected to span the moderate and severe ends of the imbalance spectrum and to allow direct connection between the grid sensitivity findings and the activation-level representations they produced.

\paragraph{ecoli3: From Activation Suppression to Feature Expansion}
Under the baseline condition, \texttt{ecoli3} produces only two active edges (\texttt{F1:mcg} and \texttt{F6:alm1}) both following monotonically increasing trajectories with narrow cross-seed variance. This is the activation signature of majority-class dominated optimization. The spline coefficients have settled into smooth, monotone mappings that track the majority-class distribution without developing the non-linear curvature that minority-class boundaries require. The narrow variance confirms that this is a stable convergent state, not an unstable failure. Yet \texttt{ecoli3} achieves $G_{\text{mean}}=0.00\%$ under the baseline condition, meaning a stable, well-converged model has learned to classify every sample as majority-class. The activation profiles make this visible as two monotone edges are insufficient boundary complexity for the seven-dimensional minority-class surface of \texttt{ecoli3}. 

Under resampling, the profile changes on both dimensions. Four edges become active (adding \texttt{F2:gvh} and \texttt{F5:aac} to the two previously active features) and the curve shapes become non-monotone. \texttt{F6:alm1} develops a concave-then-rising profile, and \texttt{F2:gvh} shows a local minimum near zero input intensity with widened cross-seed variance. This expanded, non-monotone activation set is the representation that produces $G_{\text{mean}}=87.6\%$. SMOTE-Tomek augmentation increases minority-class density across the feature space. This  provides the B-spline coefficient updates with sufficient minority observations to break out of the monotone majority attractor and develop local curvature in the minority-class region. The widened variance on \texttt{F2:gvh} indicates that the minority-class boundary in this feature dimension is genuinely complex (different random seeds find different local boundary configurations) but all seeds reach the same discriminative outcome.

\paragraph{yeast5: Feature Suppression and Recovery Under Severe Imbalance} \texttt{yeast5} ($\text{IR}=32.73$) presents a more nuanced pattern. Under the baseline condition, two edges are active and \texttt{F1:mcg} produces a bell-shaped curve with an extremely wide variance band. The band width spans the full output range, indicating that different seeds learn qualitatively different activation shapes for this feature. \texttt{F3:alm} produces a near-flat trajectory around $-0.5$ across the full input range, indicating feature suppression. The spline has converged to a near-constant mapping that contributes almost no discriminative signal. Despite these unstable and suppressed activations, \texttt{yeast5} achieves $G_{\text{mean}}=62.1\%$ at baseline (higher than \texttt{ecoli3}). This occurs because the yeast datasets are structured so that even a partially functional majority-biased classifier captures some minority-class variation through incidental correlation with majority-class features. The wide variance band on \texttt{F1:mcg} is consistent with the Phase~I finding that \texttt{yeast5}'s baseline condition maintained persistently high cross seed variance across the full grid resolution spectrum. This demonstrates that instability at the coefficient level propagates to instability in the performance trajectory.

Under resampling, two observations are structurally significant. First, \texttt{F3:alm} recovers from its flat baseline profile to develop a clearly structured non-monotone curve (rising from $-0.1$ to $+1.1$ over the central input range before plateauing). This indicates that SMOTE-Tomek augmentation provided sufficient minority-class coverage in this feature's activation domain for its spline coefficients to escape the suppressed equilibrium. This is the clearest single-feature demonstration in the study that resampling changes what the KAN learns, not merely how it weights the loss. Second, the cross-seed variance on \texttt{F1:mcg} remains wide under resampling despite the $31.2$-percentage-point gain in $G_{\text{mean}}$. This confirms the Phase~I finding that the severe underlying imbalance ($\text{IR}=32.73$) sustains optimization variability even after distributional correction. Resampling resolves the suppression of \texttt{F3:alm} and recovers mean performance, but it does not fully eliminate the initialization sensitivity introduced by extreme imbalance.

\paragraph{Interpretability as a Direct Property, Not a Post-Hoc Estimate}

The activation profiles in Fig.~\ref{fig:spline_audit} illustrate a property that has no direct equivalent in the MLP baseline. Each curve is the model's complete learned transformation for one input feature, readable as a univariate function without approximation or attribution. A domain expert examining \texttt{F3:alm}'s resampled profile can observe directly that the model assigns increasing discriminative weight to \texttt{alm} values in the central input range and a plateau above a threshold. A pattern that can be cross-referenced against domain knowledge of yeast subcellular localization without any intermediate attribution step. The MLP baseline, by contrast, encodes feature contributions through globally coupled weight matrices. It isolates the role of \texttt{alm} that would require SHAP, LIME, or gradient-based attribution, each
of which introduces approximation error and smoothing. This native interpretability is independent of classification performance. The baseline KAN's monotone \texttt{F1:mcg} curve for \texttt{ecoli3} is equally readable, and it reveals \textit{why} the model fails. It does not merely show that it fails, which the $G_{\text{mean}}$ already shows. For applications in biology, clinical diagnostics, or industrial fault detection where minority-class cases correspond to rare but consequential events, the ability to audit learned feature transformations directly is a practical advantage that the performance metrics alone cannot capture.

\subsection{Deployment Characterization}
\label{subsec:suitability_rationale}

The combined Phase~I and Phase~II findings support a characterization of KAN deployment suitability that is more precise than a simple cost benefit summary. Four structured observations frame this characterization.

\textit{The performance question is not which model wins, but under which conditions.} The non-significant performance difference between the resampled KAN and resampled MLP ($p=0.2754$, $d=+0.10$) establishes that the two architectures are statistically equivalent when the training distribution is corrected. The KAN's macro-averaged lead of $2.2$ percentage points in $G_{\text{mean}}$ is not cohort-consistent. On low-IR datasets ($\text{IR} < 12$) the MLP matches or exceeds the KAN. The KAN's advantage concentrates at IR $\geq 28$, precisely the regime where denser synthetic augmentation most expands the minority activation domain of individual spline edges. The deployment question therefore becomes: does the task involve severe imbalance where resampling is feasible? If so, the KAN achieves equivalent or superior classification performance to a matched MLP with the interpretability advantage as demonstrated in Section~\ref{subsec:manifold_audit}. If not, a simpler architecture suffices.

\textit{The cost structure is fixed, not strategy-dependent.} The KAN training overhead of $30$--$43\times$ relative to the parameter-matched MLP is consistent across all three strategies (mean ratios: $42.8\times$ baseline, $39.0\times$ focal, $43.8\times$ resampled). This reflects an architecture-level property of the current sequential spline evaluation in \texttt{pykan}, not a consequence of the training condition. This finding has a direct implication as switching strategies does not reduce KAN's computational cost. An organization deploying KAN for periodic batch retraining on imbalanced tabular data incurs the same overhead regardless of whether it uses focal loss or resampling. This makes the strategy choice a performance decision rather than a cost decision. The grid selection from Phase~I is itself a cost-efficiency decision embedded in this overhead. Selection of $G_{\text{grid}}=7$ over $G_{\text{grid}}=9$ reduces parameter count by $22\%$ while preserving the performance plateau. This displays a concrete example of where the calibration sweep recovers efficiency that would otherwise be spent on additional spline knots with diminishing returns.

\textit{The focal loss condition does not offer a cost-performance middle ground.} Focal loss produces a negligible and non-cohort-consistent improvement in KAN performance ($d=-0.17$) while incurring the same training overhead as the baseline. It therefore occupies an unfavorable position in the cost-performance space. It does not recover the training efficiency of the MLP; it does not achieve the classification performance of the resampled KAN; and it does not provide the interpretability benefit of inspecting activations shaped by a corrected training distribution. Its selective utility (improving KAN performance on moderate-IR datasets while failing on extreme-IR cases) makes it a viable supplementary tool on datasets where SMOTE-Tomek augmentation is impractical, but not a general substitute for data-level correction.

\textit{A fourth observation concerns the practical benchmark that KAN and MLP must clear to justify their deployment complexity.} XGBoost achieves $G_{\text{mean}} = 60.10\%$ under the native class distribution without any mitigation. This exceeds both the unmitigated KAN (28.64\%) and MLP (49.48\%) by substantial margins, and never failing completely on any dataset (minimum 22.36\%). This baseline robustness reflects the recursive partitioning structure of gradient-boosted trees, which inherently constructs asymmetric splits that can capture minority-class patterns under native imbalance. This further justifies a property that neither neural architecture possesses without external intervention. Under the resampled condition, the KAN surpasses the resampled XGBoost on 8 of 10 datasets, but this advantage is not statistically consistent ($p=0.0645$, $d=+0.23$). The practical implication is that XGBoost with no intervention already outperforms KAN and MLP with focal loss on this cohort, and the resampled KAN's advantage over resampled XGBoost (while real on average) is too variable across datasets to guarantee a consistent gain. For settings where interpretability is not required and training resources are constrained, XGBoost remains the most cost-efficient choice. The KAN's deployment advantage is specific to the intersection of severe imbalance, feasible resampling, and auditability requirements.

These four observations together define the boundary of KAN suitability for imbalanced tabular classification. The architecture is well-suited to settings where (i)~severe class imbalance is present, (ii)~data-level resampling is feasible, and (iii)~model auditability through direct inspection of learned feature transformations is valued. Outside these conditions, a parameter-matched MLP or XGBoost achieves comparable performance at substantially lower training cost. The findings of this study are bounded to small-to-medium scale tabular data under the current \texttt{pykan} implementation. The latency gap is partially implementation-dependent and may narrow as optimized CUDA kernels for spline evaluation become available, which would shift the cost axis of the deployment characterization without altering the performance findings.

\section{Conclusion}
\label{sec:conclusion}

This study asked whether Kolmogorov-Arnold Networks are suitable for imbalanced tabular classification. The answer is conditional and structurally grounded in the local adaptivity of KAN spline activations. This very property causes failure under native class distributions and enables strength under resampled ones. Without distributional correction, individual edge functions receive insufficient minority-class gradient signal and converge to majority-dominated mappings. This produces complete classification failure on five of ten datasets and a cohort-averaged $G_{\text{mean}}$ of $28.64\%$, more severe than the parameter-matched MLP. Focal Loss cannot resolve this, as it reweights gradients without supplying the local minority observations spline coefficients require. SMOTE-Tomek resampling by increasing minority-class density, enables spline coefficients to develop structured activation profiles. It secures the best cohort-wide rank and eliminates complete-failure cases. KANs directy inspectable spline profiles provide a native interpretability advantage unavailable from MLP weight matrices, representing independent practical value in auditable deployment settings. KAN training nonetheless requires $30$--$43\times$ more time than a parameter-matched MLP, an architecture-level cost independent of training strategy. Future work will pursue optimized spline evaluation kernels to mitigate implementation-level overhead and KAN-specific imbalance adaptations to reduce dependence on external resampling. Additionally, validation on higher-dimensional and non-tabular settings will determine whether the IR-stratified advantage observed here generalizes beyond small-to-medium tabular data.

\bibliographystyle{IEEEtran} 

\bibliography{main}

@article{kan50,
	title = {A {Generalized} {Rank}-{Order} {Method} for {Nonparametric} {Analysis} of {Data} from {Exercise} {Science}: {A} {Tutorial}},
	volume = {70},
	issn = {0270-1367},
	shorttitle = {A {Generalized} {Rank}-{Order} {Method} for {Nonparametric} {Analysis} of {Data} from {Exercise} {Science}},
	doi = {https://doi.org/10.1080/02701367.1999.10607726},
	number = {1},
	journal = {Research Quarterly for Exercise and Sport},
	publisher = {Routledge},
	author = {Thomas, Jerry R. and Nelson, Jack K. and Thomas, Katherine T.},
	year = {1999},
	pages = {11--23},
}

@article{kan27,
	title = {Geometric {Mean} {Metric} {Learning}},
	doi = {https://doi.org/10.48550/arXiv.1607.05002},
	journal = {arXiv:1607.05002},
	author = {Zadeh, Pourya Habib and Hosseini, Reshad and Sra, Suvrit},
	year = {2016},
}

@article{kan16,
	title = {{BSRBF}-{KAN}: {A} combination of {B}-splines and {Radial} {Basis} {Functions} in {Kolmogorov}-{Arnold} {Networks}},
	copyright = {Creative Commons Attribution Non Commercial Share Alike 4.0 International},
	shorttitle = {{BSRBF}-{KAN}},
	url = {https://arxiv.org/abs/2406.11173},
	doi = {https://doi.org/10.48550/ARXIV.2406.11173},
	urldate = {2026-06-28},
	journal = {ARXIV.2406.11173},
	author = {Ta, Hoang-Thang},
	year = {2024},
}

@inproceedings{kan8,
	title = {A {Benchmarking} {Study} of {Kolmogorov}-{Arnold} {Networks} on {Tabular} {Data}},
	doi = {https://doi.org/10.1109/AICT61888.2024.10740444},
	booktitle = {{AICT}},
	publisher = {IEEE Xplore},
	author = {Poeta, Eleonora and Giobergia, Flavio and Pastor, Eliana and Cerquitelli, Tania and Baralis, Elena},
	year = {2024},
	pages = {1--6},
}

@inproceedings{kan44,
	title = {Prediction of {Muscle} {Power} in {Elderly} {Using} {Functional} {Screening} {Data}},
	issn = {2770-4319},
	doi = {https://doi.org/10.1109/MIPR59079.2023.00017},
	booktitle = {{IEEE} {MIPR}},
	author = {Vijay, Vivek and Shukla, Brajesh Kumar and Yadav, Sandeep Kumar and Yadav, Pankaj},
	year = {2023},
	pages = {1--6},
}

@article{kan40,
	title = {Rebalancing the {Scales}: {A} {Systematic} {Mapping} {Study} of {Generative} {Adversarial} {Networks} ({GANs}) in {Addressing} {Data} {Imbalance}},
	shorttitle = {Rebalancing the {Scales}},
	journal = {arXiv.2502.16535},
	author = {Yadav, Pankaj and Sihag, Gulshan and Vijay, Vivek},
	year = {2025},
	note = {arXiv:2502.16535 [cs.AI]},
}

@article{kan19,
	title = {On the {Convergence} of ({Stochastic}) {Gradient} {Descent} for {Kolmogorov}--{Arnold} {Networks}},
	doi = {https://doi.org/10.48550/arXiv.2410.08041},
	journal = {arXiv.2410.08041},
	author = {Gao, Yihang and Tan, Vincent Y. F.},
	year = {2024},
}

@article{kan6,
	title = {{KAN} or {MLP}: {A} {Fairer} {Comparison}},
	shorttitle = {{KAN} or {MLP}},
	doi = {https://doi.org/10.48550/arXiv.2407.16674},
	journal = {arXiv.2407.16674},
	author = {Yu, Runpeng and Yu, Weihao and Wang, Xinchao},
	year = {2024},
}

@article{kan7,
	title = {Physics {Informed} {Kolmogorov}-{Arnold} {Neural} {Networks} for {Dynamical} {Analysis} via {Efficent}-{KAN} and {WAV}-{KAN}},
	doi = {https://doi.org/10.48550/arXiv.2407.18373},
	journal = {arXiv.2407.18373},
	author = {Patra, Subhajit and Panda, Sonali and Parida, Bikram Keshari and Arya, Mahima and Jacobs, Kurt and Bondar, Denys I. and Sen, Abhijit},
	year = {2024},
}

@article{kan11,
	title = {The {Kolmogorov} {Superposition} {Theorem} can {Break} the {Curse} of {Dimensionality} {When} {Approximating} {High} {Dimensional} {Functions}},
	doi = {https://doi.org/10.48550/arXiv.2112.09963},
	journal = {arXiv.2112.09963},
	author = {Lai, Ming-Jun and Shen, Zhaiming},
	year = {2024},
}

@article{kan13,
	title = {{rKAN}: {Rational} {Kolmogorov}-{Arnold} {Networks}},
	shorttitle = {{rKAN}},
	doi = {https://doi.org/10.48550/arXiv.2406.14495},
	journal = {arXiv.2406.14495},
	author = {Aghaei, Alireza Afzal},
	year = {2024},
}

@article{kan18,
	title = {{PRKAN}: {Parameter}-{Reduced} {Kolmogorov}-{Arnold} {Networks}},
	shorttitle = {{PRKAN}},
	doi = {https://doi.org/10.48550/arXiv.2501.07032},
	journal = {arXiv.2501.07032},
	author = {Ta, Hoang-Thang and Thai, Duy-Quy and Tran, Anh and Sidorov, Grigori and Gelbukh, Alexander},
	year = {2025},
}

@article{kan17,
	title = {A {Comprehensive} {Survey} on {Kolmogorov} {Arnold} {Networks} ({KAN})},
	doi = {https://doi.org/10.48550/arXiv.2407.11075},
	journal = {arXiv:2407.11075},
	author = {Ji, Tianrui and Hou, Yuntian and Zhang, Di},
	year = {2025},
}

@article{kan4,
	title = {{KAN}: {Kolmogorov}-{Arnold} {Networks}},
	shorttitle = {{KAN}},
	doi = {https://doi.org/10.48550/arXiv.2404.19756},
	journal = {arXiv:2404.19756},
	author = {Liu, Ziming and Wang, Yixuan and Vaidya, Sachin and Ruehle, Fabian and Halverson, James and Soljačić, Marin and Hou, Thomas Y. and Tegmark, Max},
	year = {2025},
}

@article{kan22,
	title = {Enhancing {Synthetic} {Oversampling} for {Imbalanced} {Datasets} {Using} {Proxima}-{Orion} {Neighbors} and q-{Gaussian} {Weighting} {Technique}},
	doi = {https://doi.org/10.48550/arXiv.2501.15790},
	journal = {arXiv.2501.15790},
	author = {Yadav, Pankaj and Vijay, Vivek and Sihag, Gulshan},
	year = {2025},
}

@article{kan52,
	title = {Underpowered samples, false negatives, and unconscious learning},
	volume = {23},
	issn = {1531-5320},
	doi = {https://doi.org/10.3758/s13423-015-0892-6},
	language = {en},
	number = {1},
	journal = {Psychonomic Bulletin \& Review},
	author = {Vadillo, Miguel A. and Konstantinidis, Emmanouil and Shanks, David R.},
	year = {2016},
	pages = {87--102},
}

@article{kan51,
	title = {Basic statistical tools in research and data analysis},
	volume = {60},
	issn = {0019-5049},
	doi = {https://doi.org/10.4103/0019-5049.190623},
	language = {en-US},
	number = {9},
	journal = {Indian Journal of Anaesthesia},
	author = {Ali, Zulfiqar and Bhaskar, S. Bala},
	year = {2016},
	pages = {662},
}

@article{kan49,
	title = {Emerging {SMOTE} and {GAN} {Variants} for {Data} {Augmentation} in {Imbalance} {Machine} {Learning} {Tasks}: {A} {Review}},
	volume = {13},
	issn = {2169-3536},
	shorttitle = {Emerging {SMOTE} and {GAN} {Variants} for {Data} {Augmentation} in {Imbalance} {Machine} {Learning} {Tasks}},
	doi = {https://doi.org/10.1109/ACCESS.2025.3584532},
	journal = {IEEE Access},
	publisher = {IEEE Xplore},
	author = {Udu, Amadi G. and Salman, Marwah T. and Ghalati, Maryam K. and Lecchini-Visintini, Andrea and Siddle, David R. and Dong, Hongbiao},
	year = {2025},
	pages = {113838--113853},
}

@article{kan48,
	title = {Automatically weighted focal loss for imbalance learning},
	volume = {37},
	issn = {1433-3058},
	doi = {https://doi.org/10.1007/s00521-024-10323-x},
	language = {en},
	number = {5},
	journal = {Neural Computing and Applications},
	publisher = {Springer Link},
	author = {Mahmoodi, Nasibeh and Shirazi, Hossein and Fakhredanesh, Mohammad and DadashtabarAhmadi, Koroush},
	year = {2025},
	pages = {4035--4052},
}

@article{kan47,
	title = {Bayesian {Kolmogorov}-{Arnold} networks: {Uncertainty}-aware interpretable modeling through probabilistic spline decomposition},
	volume = {680},
	issn = {0378-4371},
	shorttitle = {Bayesian {Kolmogorov}-{Arnold} networks},
	doi = {https://doi.org/10.1016/j.physa.2025.131041},
	journal = {Physica A: Statistical Mechanics and its Applications},
	publisher = {ScienceDirect},
	author = {Hassan, Masoud Muhammed},
	year = {2025},
	pages = {131041},
}

@article{kan46,
	title = {Imbalance-{XGBoost}: leveraging weighted and focal losses for binary label-imbalanced classification with {XGBoost}},
	volume = {136},
	issn = {0167-8655},
	shorttitle = {Imbalance-{XGBoost}},
	doi = {https://doi.org/10.1016/j.patrec.2020.05.035},
	journal = {Pattern Recognition Letters},
	publisher = {ScienceDirect},
	author = {Wang, Chen and Deng, Chengyuan and Wang, Suzhen},
	year = {2020},
	pages = {190--197},
}

@article{kan45,
	title = {{FC}-{KAN}: {Function} combinations in {Kolmogorov}-{Arnold} networks},
	volume = {736},
	issn = {0020-0255},
	shorttitle = {{FC}-{KAN}},
	doi = {https://doi.org/10.1016/j.ins.2026.123103},
	journal = {Information Sciences},
	publisher = {ScienceDirect},
	author = {Ta, Hoang-Thang and Thai, Duy-Quy and Rahman, Abu Bakar Siddiqur and Sidorov, Grigori and Gelbukh, Alexander},
	year = {2026},
	pages = {123103},
}

@article{kan43,
	title = {The class imbalance problem in deep learning},
	volume = {113},
	issn = {1573-0565},
	doi = {https://doi.org/10.1007/s10994-022-06268-8},
	language = {en},
	number = {7},
	journal = {Machine Learning},
	publisher = {Springer Link},
	author = {Ghosh, Kushankur and Bellinger, Colin and Corizzo, Roberto and Branco, Paula and Krawczyk, Bartosz and Japkowicz, Nathalie},
	year = {2024},
	pages = {4845--4901},
}

@article{kan41,
	title = {A comprehensive review on data-level methods for imbalanced data classification},
	volume = {295},
	issn = {0957-4174},
	doi = {10.1016/j.eswa.2025.128920https://doi.org/},
	journal = {Expert Systems with Applications},
	publisher = {ScienceDirect},
	author = {Nikpour, Bahareh and Rahmati, Farshad and Mirzaei, Behzad and Nezamabadi-pour, Hossein},
	year = {2026},
	pages = {128920},
}

@article{kan39,
	title = {{KANBalance}: {Kolmogorov}–{Arnold} network mitigates class imbalance},
	volume = {171},
	issn = {0031-3203},
	shorttitle = {{KANBalance}},
	doi = {https://doi.org/10.1016/j.patcog.2025.112325},
	journal = {Pattern Recognition},
	publisher = {ScienceDirect},
	author = {Qezelbash-Chamak, Jaber and Hicklin, Karen and Kim, Minhee},
	year = {2026},
	pages = {112325},
}

@article{kan37,
	title = {{MLSTL}-{WSN}: machine learning-based intrusion detection using {SMOTETomek} in {WSNs}},
	volume = {23},
	issn = {1615-5270},
	shorttitle = {{MLSTL}-{WSN}},
	doi = {https://doi.org/10.1007/s10207-024-00833-z},
	language = {en},
	number = {3},
	journal = {International Journal of Information Security},
	publisher = {Springer Link},
	author = {Talukder, Md. Alamin and Sharmin, Selina and Uddin, Md Ashraf and Islam, Md Manowarul and Aryal, Sunil},
	year = {2024},
	pages = {2139--2158},
}

@article{kan36,
	title = {Representation {Learning} for {Tabular} {Data}: {A} {Comprehensive} {Survey}},
	volume = {48},
	issn = {1939-3539},
	shorttitle = {Representation {Learning} for {Tabular} {Data}},
	doi = {https://doi.org/10.1109/TPAMI.2026.3657217},
	number = {6},
	journal = {IEEE Transactions on Pattern Analysis and Machine Intelligence},
	publisher = {IEEE Xplore},
	author = {Jiang, Jun-Peng and Liu, Si-Yang and Cai, Hao-Run and Zhou, Qi-Le and Ye, Han-Jia},
	year = {2026},
	pages = {6488--6508},
}

@article{kan34,
	title = {Augmented {Intelligence} of {Things} for {Emergency} {Vehicle} {Secure} {Trajectory} {Prediction} and {Task} {Offloading}},
	volume = {11},
	issn = {2327-4662},
	doi = {https://doi.org/10.1109/JIOT.2024.3417285},
	number = {22},
	journal = {IEEE Internet of Things Journal},
	publisher = {IEEE Xplore},
	author = {Wu, Xiang and Dong, Jian and Bao, Wei and Zou, Baowen and Wang, Lili and Wang, Huanhuan},
	year = {2024},
	pages = {36030--36043},
}

@article{kan35,
	title = {Dynamic {Security} {Computing} {Framework} {With} {Zero} {Trust} {Based} on {Privacy} {Domain} {Prevention} and {Control} {Theory}},
	volume = {43},
	issn = {1558-0008},
	doi = {https://doi.org/10.1109/JSAC.2025.3560036},
	number = {6},
	journal = {IEEE Journal on Selected Areas in Communications},
	publisher = {IEEE Xplore},
	author = {Wu, Xiang and Zou, Baowen and Lu, Chuanchuan and Wang, Lili and Zhang, Yongting and Wang, Huanhuan},
	year = {2025},
	pages = {2266--2278},
}

@inproceedings{kan33,
	title = {Self-{Supervised} {Learning} on {A} {Lightweight} {Low}-{Light} {Image} {Enhancement} {Model} with {Curve} {Refinement}},
	issn = {2379-190X},
	doi = {https://doi.org/10.1109/ICASSP43922.2022.9746348},
	booktitle = {{ICASS}},
	publisher = {IEEE Xplore},
	author = {Wu, Wanyu and Wang, Wei and Jiang, Kui and Xu, Xin and Hu, Ruimin},
	year = {2022},
	pages = {1890--1894},
}

@article{kan32,
	title = {Low-{Light} {Salient} {Object} {Detection} {Meets} the {Small} {Size}},
	volume = {9},
	issn = {2471-285X},
	doi = {https://doi.org/10.1109/TETCI.2024.3442819},
	number = {4},
	journal = {IEEE Transactions on Emerging Topics in Computational Intelligence},
	publisher = {IEEE Transactions on Emerging Topics in Computational Intelligence},
	author = {Wang, Shiqin and Xu, Xin and Chen, Haoyang and Jiang, Kui and Wang, Zheng and Tang, Ke},
	year = {2025},
	pages = {2754--2766},
}

@inproceedings{kan31,
	address = {New York, NY, USA},
	series = {{MM} '23},
	title = {Informative {Classes} {Matter}: {Towards} {Unsupervised} {Domain} {Adaptive} {Nighttime} {Semantic} {Segmentation}},
	shorttitle = {Informative {Classes} {Matter}},
	doi = {https://doi.org/10.1145/3581783.3611956},
	booktitle = {31st {ACM}-{MM}},
	publisher = {Association for Computing Machinery},
	author = {Wang, Shiqin and Xu, Xin and Ma, Xianzheng and Jiang, Kui and Wang, Zheng},
	year = {2023},
	pages = {163--172},
}

@article{kan30,
	title = {Robustness and resilience of complex networks},
	volume = {6},
	issn = {2522-5820},
	doi = {https://doi.org/10.1038/s42254-023-00676-y},
	language = {en},
	number = {2},
	journal = {Nature Reviews Physics},
	author = {Artime, Oriol and Grassia, Marco and De Domenico, Manlio and Gleeson, James P. and Makse, Hernán A. and Mangioni, Giuseppe and Perc, Matjaž and Radicchi, Filippo},
	year = {2024},
	pages = {114--131},
}

@article{kan29,
	series = {Signal propagation in complex networks},
	title = {Signal propagation in complex networks},
	volume = {1017},
	issn = {0370-1573},
	doi = {https://doi.org/10.1016/j.physrep.2023.03.005},
	journal = {Physics Reports},
	author = {Ji, Peng and Ye, Jiachen and Mu, Yu and Lin, Wei and Tian, Yang and Hens, Chittaranjan and Perc, Matjaž and Tang, Yang and Sun, Jie and Kurths, Jürgen},
	year = {2023},
	pages = {1--96},
}

@article{kan24,
	title = {Evaluation metrics and statistical tests for machine learning},
	volume = {14},
	issn = {2045-2322},
	doi = {https://doi.org/10.1038/s41598-024-56706-x},
	language = {eng},
	number = {1},
	journal = {Scientific Reports},
	publisher = {Scientific Reports},
	author = {Rainio, Oona and Teuho, Jarmo and Klén, Riku},
	year = {2024},
	pages = {6086},
}

@article{kan21,
	title = {{KEEL} {Data}-{Mining} {Software} {Tool}: {Data} {Set} {Repository}, {Integration} of {Algorithms} and {Experimental} {Analysis} {Framework}},
	volume = {13},
	url = {https://doi: 10.1007/s00500-008-0323-y},
	language = {en},
	number = {3},
	journal = {Soft Computing},
	author = {Alcalá-Fdez, J and Fernández, A and Luengo, J and Derrac, J and García, S and Sánchez, L and Herrera, F},
	year = {2009},
	pages = {307--318},
}

@article{kan20,
	title = {Synergetic {Focal} {Loss} for {Imbalanced} {Classification} in {Federated} {XGBoost}},
	volume = {5},
	issn = {2691-4581},
	url = {https://ieeexplore.ieee.org/abstract/document/10063985},
	doi = {https://doi.org/10.1109/TAI.2023.3254519},
	number = {2},
	urldate = {2025-07-09},
	journal = {IEEE Transactions on Artificial Intelligence},
	publisher = {IEEE Xplore},
	author = {Tian, Jiao and Tsai, Pei-Wei and Zhang, Kai and Cai, Xinyi and Xiao, Hongwang and Yu, Ke and Zhao, Wenyu and Chen, Jinjun},
	year = {2024},
	pages = {647--660},
}

@article{kan15,
	title = {The {Numerical} {Evaluation} of {B}-{Splines}*},
	volume = {10},
	issn = {0272-4960},
	doi = {https://doi.org/10.1093/imamat/10.2.134},
	number = {2},
	journal = {IMA Journal of Applied Mathematics},
	publisher = {Academic Press Inc. (London) Limited},
	author = {COX, M. G.},
	year = {1972},
	pages = {134--149},
}

@article{kan14,
	title = {C-{Bézier} {Curves} and {Surfaces}},
	volume = {61},
	issn = {1077-3169},
	doi = {https://doi.org/10.1006/gmip.1999.0490},
	number = {1},
	journal = {Graphical Models and Image Processing},
	publisher = {ScienceDirect},
	author = {Zhang, Jiwen},
	year = {1999},
	pages = {2--15},
}

@book{kan12,
	title = {Mathematical problems},
	doi = {https://doi.org/10.13140/RG.2.2.10904.81922},
	language = {en},
	publisher = {Chapman and Hall/CRC},
	author = {Hilbert, D.},
	year = {2019},
}

@inproceedings{kan10,
	title = {Kolmogorov {Superposition} {Theorem} and {Its} {Application} to {Multivariate} {Function} {Decompositions} and {Image} {Representation}},
	doi = {https://doi.org/10.1109/SITIS.2008.16},
	booktitle = {{SITIS}},
	publisher = {IEEE Xplore},
	author = {Leni, Pierre-Emmanuel and Fougerolle, Yohan D. and Truchetet, Frédéric},
	year = {2008},
	pages = {344--351},
}

@article{kan9,
	title = {A {Survey} on {Kolmogorov}-{Arnold} {Network}},
	issn = {0360-0300},
	doi = {https://doi.org/10.1145/3743128},
	journal = {ACM Comput. Surv.},
	publisher = {ACM Digital Library},
	author = {Somvanshi, Shriyank and Javed, Syed Aaqib and Islam, Md Monzurul and Pandit, Diwas and Das, Subasish},
	year = {2025},
}

@article{kan5,
	title = {A universal mapping for kolmogorov's superposition theorem},
	volume = {6},
	issn = {0893-6080},
	doi = {https://doi.org/10.1016/S0893-6080(09)80020-8},
	number = {8},
	journal = {Neural Networks},
	publisher = {Neural Networks},
	author = {Sprecher, David A.},
	year = {1993},
	pages = {1089--1094},
}

@inproceedings{kan3,
	title = {Neural {Networks} {Architectures} {Design}, and {Applications}: {A} {Review}},
	shorttitle = {Neural {Networks} {Architectures} {Design}, and {Applications}},
	doi = {https://doi.org/10.1109/ICOASE51841.2020.9436582},
	booktitle = {2020 {International} {Conference} on {Advanced} {Science} and {Engineering} ({ICOASE})},
	publisher = {IEEE Xplore},
	author = {Sadeeq, Mohammed A.M. and Abdulazeez, Adnan M.},
	year = {2020},
	pages = {199--204},
}

@article{kan2,
	title = {Performance of evaluation metrics for classification in imbalanced data},
	volume = {40},
	issn = {1613-9658},
	url = {https://doi.org/10.1007/s00180-024-01539-5},
	doi = {10.1007/s00180-024-01539-5},
	language = {en},
	number = {3},
	journal = {Computational Statistics},
	author = {de la Cruz Huayanay, Alex and Bazán, Jorge L. and Russo, Cibele M.},
	year = {2025},
	pages = {1447--1473},
}

@inproceedings{kan42,
	title = {Evaluation of {Risk} {Factors} for {Fall} in {Elderly} {People} from {Imbalanced} {Data} using the {Oversampling} {Technique} {SMOTE}:},
	isbn = {978-989-758-566-1},
	shorttitle = {Evaluation of {Risk} {Factors} for {Fall} in {Elderly} {People} from {Imbalanced} {Data} using the {Oversampling} {Technique} {SMOTE}},
	doi = {https://doi.org/10.5220/0011041200003188},
	language = {en},
	booktitle = {8th {ICT4AWE}},
	publisher = {SCITEPRESS},
	author = {Sihag, Gulshan and Yadav, Pankaj and Delcroix, Veronique and Vijay, Vivek and Siebert, Xavier and Yadav, Sandeep and Puisieux, François},
	year = {2022},
	pages = {50--58},
}

@article{kan38,
	title = {Explainable {AI} for {Intrusion} {Detection} {Systems}: {LIME} and {SHAP} {Applicability} on {Multi}-{Layer} {Perceptron}},
	volume = {12},
	issn = {2169-3536},
	shorttitle = {Explainable {AI} for {Intrusion} {Detection} {Systems}},
	doi = {https://doi.org/10.1109/ACCESS.2024.3368377},
	journal = {IEEE Access},
	author = {Gaspar, Diogo and Silva, Paulo and Silva, Catarina},
	year = {2024},
	pages = {30164--30175},
}

@article{kan1,
	title = {Imbalanced {Data} {Problem} in {Machine} {Learning}: {A} {Review}},
	volume = {13},
	issn = {2169-3536},
	shorttitle = {Imbalanced {Data} {Problem} in {Machine} {Learning}},
	doi = {https://doi.org/10.1109/ACCESS.2025.3531662},
	journal = {IEEE Access},
	author = {Altalhan, Manahel and Algarni, Abdulmohsen and Turki-Hadj Alouane, Monia},
	year = {2025},
	pages = {13686--13699},
}

\vfill

\end{document}